\documentclass[lettersize,journal]{IEEEtran}
\usepackage{amsmath,amsfonts}
\usepackage{algorithmic}
\usepackage{algorithm}
\usepackage{array}
\usepackage[caption=false,font=normalsize,labelfont=sf,textfont=sf]{subfig}
\usepackage{textcomp}
\usepackage{stfloats}
\usepackage{url}
\usepackage{verbatim}
\usepackage{booktabs}
\usepackage{threeparttable}
\usepackage{array}
\usepackage{tabularx}
\usepackage{multirow}
\usepackage{float}
\usepackage{graphicx}
\usepackage{cite}
\newcommand{\cmark}{\(\checkmark\)}
\newcommand{\pmark}{\(\triangle\)}
\newcommand{\xmark}{\(\times\)}
\newcommand{\vhead}[1]{\raisebox{1ex}{\textbf{#1}}}

\newcolumntype{L}[1]{>{\raggedright\arraybackslash}p{#1}}
\newcolumntype{C}[1]{>{\centering\arraybackslash}p{#1}}
\newcolumntype{Y}{>{\centering\arraybackslash}X}

\begin{document}

\title{World Narrative Model for Highly Controllable Video Generation: A Paradigm Shift from Pixel Sampling to Physical World Orchestration}

\author{The World Narrative Model Team
\thanks{The World Narrative Model team members: Ye Chen, Xuanhong Chen, Yupeng Zhu, Liming Tan, Zhewen Wan, Yuxuan Xiong, Tielong Wang, Jinfan Liu, Wuze Zhang, Xiongzhen Zhang, Feifei Li, Xianglin Luo, Zhehan Zhao, Zhifan Zhang, Laisheng Kou, Zhujin Liang, Yugang Chen, Muchun Chen, Xu Miao, Yijing Zhang,  Xiaojie Sheng, Qiang Hu, Jialiang Chen, Weimin Zhang, Wenjun Zhang, Bingbing Ni.}
\thanks{The overall team is from Shanghai Jiao Tong University, and Dr. Xu Miao is from datacanvas.com.}
\thanks{Corresponding author: Dr. Bingbing Ni (nibingbing@sjtu.edu.cn) initializes and leads the project.}
}



\maketitle

\begin{abstract}
The fundamental obstacle to industrial grade video generation is the lack of controllability: existing models treat video as a pixel distribution sampling problem, bypassing the explicit, instance level $4D$ $(3D + T)$ physical world. Consequently, content creators cannot specify geometry, motion, camera parameters, or lighting in a deterministic, quantitative way, leading to the infamous "gacha" loop that makes professional content creation prohibitively inefficient and expensive.
To address this, we introduce the \emph{World Narrative Model (WNM)}, a new paradigm that fundamentally decouples what to render (structured physical narrative) from how to render (pixel generation). The WNM is the first framework that, for media generation, replaces end-to-end black-box sampling with an orchestrated $(3D + T)$ pre-visualization. It consists of a set of collaborative agents that automatically translate sparse multi-modal user inputs (text, reference video, sketches) into a fully editable, instance level world representation, including scene geometry, object placements, character/animal skeleton motions and trajectories, camera motion, and lighting parameters, all at quantitative, physically meaningful granularity. This world representation serves as a structural deterministic "blueprint" that can then drive any existing video foundation model (frozen or with lightweight adapters) to render the final footage, turning the base model into a faithful "neural shader".
On top of this engine, we build a human-AI collaborative video creation platform that not only offers automatic world generation and pre-visualization aligned with professional film-making pipelines, but also provides director manipulation consoles for scene, asset, motion, camera, and lighting, enabling seamless human refinement.
Extensive experiments show that our paradigm greatly eliminates probabilistic "gacha" calls, and produces videos whose spatial layout, character motion, and cinematography precisely follow the creator's intent. Our framework establishes an open, modular architecture where each component (\emph{e.g.}, world representation, control agents, adapter interfaces) can be independently improved by future research. Our project website: https://glassroom.sjtu.edu.cn/WNM/.

\end{abstract}


\section{Introduction}
\IEEEPARstart{R}{ecent} years have witnessed remarkable advances in video generation. Models such as Sora~\cite{liu2024sora}, Kling~\cite{klingteam2026motioncontrol}, Seedance~\cite{gao2025seedance, teamseedance2025seedance15pro, teamseedance2026seedance2}, and Wan~\cite{teamwan2025wan} produce stunning image quality, achieving $1080$ to $4K$ resolution and multi-second temporal coherence. Yet beneath this technical triumph lies a deep and widely ignored crisis: \emph{these models are practically unusable for professional content creation}.
The core issue is known as \textbf{uncontrollability}: a fundamental bottleneck for AIGC industrial. In practice, content creators face three manifestations of uncontrollability: 1) \textbf{Non-determinism}: the same prompt yields different results every time. There is no way to lock a specific output; 2) \textbf{No physical-level specification}: users cannot directly set joint angles, object poses, light positions, or exact camera trajectories. All such parameters remain hidden inside the black-box model; 3) \textbf{Infeasible local editing}: a change intended for a single object (\emph{e.g.}, raising a character's arm) often causes global drift-background textures shift, other objects deform, lighting changes unpredictably.

The control methods offered by current models are all severely limited:  
1) \emph{Text prompts} are intuitive but inherently sparse, and they cannot describe precise $3D$ geometry, material properties, or exact motion curves.  
2) \emph{Reference/anchor images} provide visual constraints but leave the intermediate motion uncontrolled, suffering from highly temporal sparsity.  
3) \emph{Motion trajectories/kinematics} give explicit motion (joint motion) paths, yet still offer no control over object identity, lighting, or articulation.  
4) \emph{Conditional injection techniques (semantic mask, depth, edge, pose)} operate only at the pixel-level; they do not expose high-level physical semantics such as "light color temperature" or "elbow joint angle".
Consequently, the industry has been forced into a "gacha" (card-drawing) production model. Data from leading short-drama production houses show that, on average, $20–50$ generations are required to obtain a single usable shot. For premium content requiring precise direction, this number can exceed $100$ trials per shot, which implies a paradox, \emph{i.e.}, the token cost of generating low-quality, uncontrollable outputs often exceeds the value of the final video product, especially in competitive markets where only high-quality content retains value.
The above defeat transforms the director's role from an orchestrator of artistic vision into a passive "prompt engineer" struggling against a stochastic black box. 

\begin{figure*}[!t]
\centering
\includegraphics[width=0.95\linewidth]{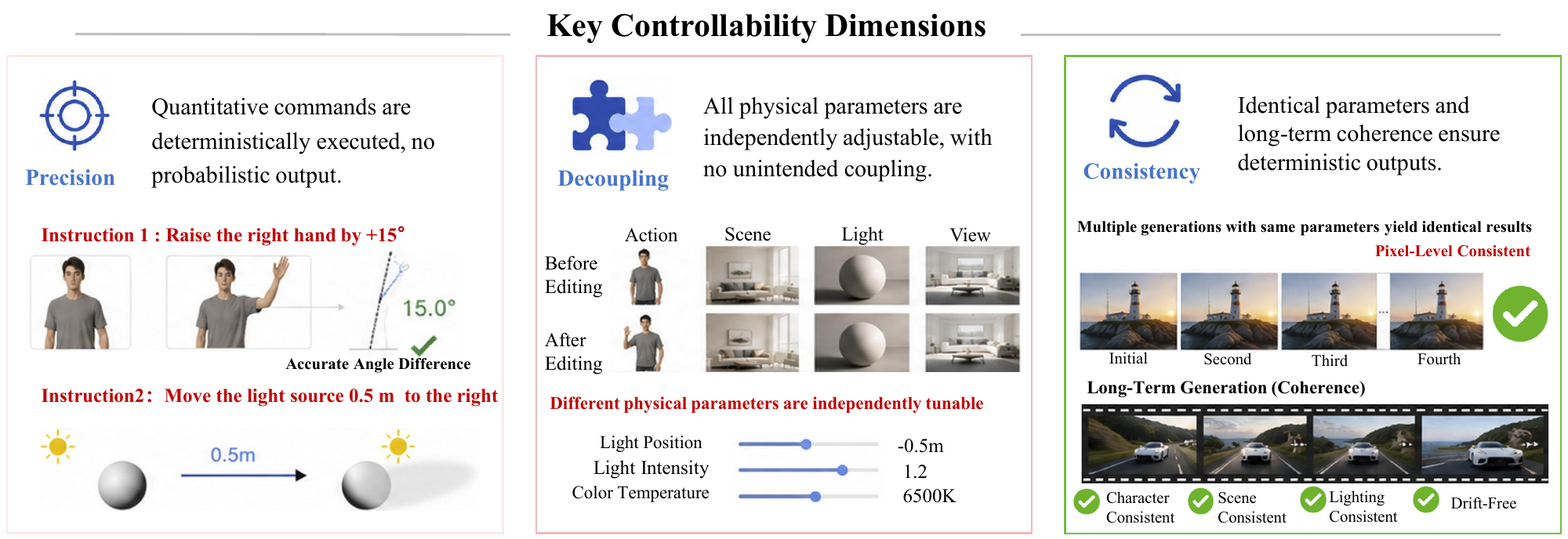}
\caption{Three key dimensions that indicate the controllability in video generation task, including \textbf{decoupling}, \textbf{consistency} and \textbf{spatial-temporal completeness}.}
\label{fig:keyindicator}
\end{figure*}

We argue that true controllability consists of the following four orthogonal dimensions, as iilustrated in Figure~\ref{fig:keyindicator}.
\begin{enumerate}
\item \emph{Precision}: the ability to specify a desired attribute (\emph{e.g.}, "right elbow flexed $15$ degrees") and have the output exactly match that value, not an descriptive approximation. This requires a \emph{quantitative}, not just qualitative, control channel.

\item \emph{Decoupling}: the ability to modify one attribute (\emph{e.g.}, character pose) or one entity without affecting unrelated attributes (\emph{e.g.}, background lighting) or other entities. Current models tightly entangle all physical properties; a local change often propagates unpredictably.

\item \emph{Consistency}: the ability to maintain the same physical state over time and across multiple trials. Given the same control parameters, the model must produce \emph{identical results} each time, \emph{i.e.}, determinism is a prerequisite for any professional production pipeline.

\item \emph{Spatio-temporal Completeness}: control must be exercisable at any point in space and time, \emph{i.e.}, specify a pose at frame $10$, a camera angle at frame $25$, a light movement over $5$ seconds. Sparse controls (\emph{e.g.}, only first and last frame) are insufficient.
\end{enumerate}

These four dimensions together define \textbf{full physical semantic controllability}: the creator can directly, quantitatively, and independently manipulate every physically meaningful parameter of the scene, at any frame, with deterministic outcome.
Unfortunately, as illustrated in Table~\ref{tab:controlcompare}, no existing video generation model or integrated video generation platform achieves controllability in any of these dimensions. State-of-the-art models (Sora, Kling, Wan, Seedance \emph{etc.}) and popular storyboard tools (TapNow~\cite{tapnow2026}, Lovart~\cite{lovart2026}, Seko~\cite{sensetime2026seko}, CapCut~\cite{capcut2026}, 360Nano~\cite{nanoai2026}, libTV~\cite{liblibai2026libtv} \emph{etc.}) offer only \emph{coarse-grained, stochastic, and tightly coupled control}, which are unsuitable for professional content creation where every single instance along with its versatile physical attributes at each spatio-temporal location must be predictable and editable. 
\begin{table*}[t]
    \centering
    \scriptsize
    \setlength{\tabcolsep}{2.4pt}
    \renewcommand{\arraystretch}{1.10}
    \begin{threeparttable}
        \caption{Comparison of control capabilities across representative video generation models and platforms. Existing systems mainly provide qualitative, prompt-driven, or reference-driven control, whereas the proposed WNM enables explicit, parameterized, decoupled, and structural deterministic control.}
        \begin{tabularx}{\textwidth}{@{}L{3cm}YYYY!{\vrule width 0.35pt}YYY!{\vrule width 0.35pt}C{2.10cm}@{}}
            \toprule
            \vhead{Control Modality} & \vhead{Seedance2.0} & \vhead{KlingO3} & \vhead{HappyHorse} & \vhead{Wan 2.7} & \vhead{TapNow} & \vhead{LibTV} & \vhead{RHTV} & \textbf{\shortstack{WNM\\(Ours)}} \\
            \midrule
            Text Prompt & \cmark & \cmark & \cmark & \cmark & \cmark & \cmark & \cmark & \cmark \\
            Reference Image & \cmark & \cmark & \cmark & \cmark & \cmark & \cmark & \cmark & \cmark \\
            First-to-Last & \cmark & \cmark & \xmark & \cmark & \cmark & \cmark & \cmark & \cmark \\
            Reference Video & \cmark & \cmark & \cmark & \cmark & \cmark & \cmark & \cmark & \cmark \\
            Temporal Keyframes & \xmark & \xmark & \xmark & \xmark & \pmark & \pmark & \pmark & \textbf{Frame-level} \\
            Omni-Control & \cmark & \cmark & \xmark & \cmark & \cmark & \cmark & \cmark & \cmark \\
            Camera Control & \xmark & \xmark & \xmark & \xmark & \pmark & \pmark & \pmark & \textbf{Parametric} \\
            Lighting Control & \xmark & \xmark & \xmark & \xmark & \pmark & \pmark & \pmark & \textbf{Parametric} \\
            Scene Layout & \xmark & \xmark & \xmark & \xmark & \pmark & \pmark & \pmark & \textbf{Parametric} \\
            Pose Control & \pmark & \pmark & \xmark & \xmark & \pmark & \pmark & \pmark & \textbf{Parametric} \\
            Character Staging & \xmark & \xmark & \xmark & \xmark & \pmark & \pmark & \pmark & \textbf{Parametric} \\
            Object-level Editing & \xmark & \xmark & \xmark & \xmark & \xmark & \xmark & \xmark & \textbf{Instance-level} \\
            Decoupled Editing & \xmark & \xmark & \xmark & \xmark & \xmark & \xmark & \xmark & \textbf{Decoupled} \\
            Deterministic Output & \xmark & \xmark & \xmark & \xmark & \xmark & \xmark & \xmark & \textbf{Deterministic} \\
            \bottomrule
        \end{tabularx}
        \begin{tablenotes}[flushleft]
            \scriptsize
            \item \cmark\ Supported. \hspace{0.8em} \pmark\ Partially or implicitly supported. \hspace{0.8em} \xmark\ Not supported.
        \end{tablenotes}
    \end{threeparttable}
    \label{tab:controlcompare}
\end{table*}

\begin{figure*}[!t]
\centering
\includegraphics[width=0.95\linewidth]{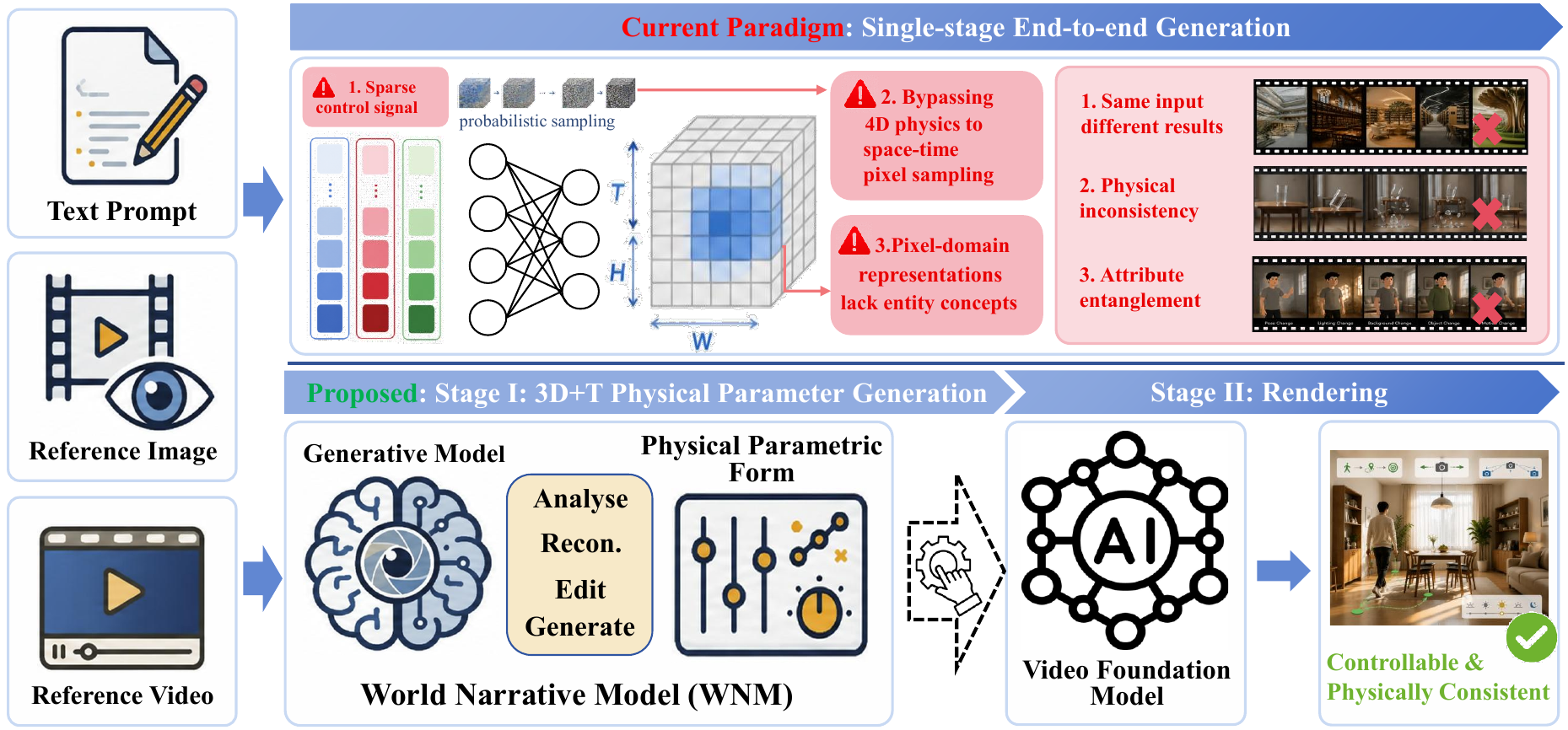}
\caption{Motivations of the proposed new video generation paradigm: from end-to-end generation towards two-phase task decoupling, \emph{i.e.}, \textbf{full-dimensional world control} and \textbf{pixel render}.}
\label{fig:motivation}
\end{figure*}

\textbf{The Essence of Uncontrollability}. All mainstream video generation foundation models are essentially high-dimensional conditional samplers built upon extremely deep, multi-layer architectures such as transformer~\cite{vaswani2017attention}. These architectures implicitly encode visual-semantic patterns from training data into a soft key-value association network through massive parameters, enabling approximate information retrieval, fusion, and response conditioning during inference. However, its internal representation does not explicitly store any physical entities or rules; instead, it treats the pixel distribution of each frame as a probability density function in an extremely high-dimensional space. On this basis, diffusion models~\cite{ho2020denoising} further introduce a stochastic iterative de-noising sampling process, making the output essentially a random walk in this high-dimensional probability space.
As illustrated in Figure~\ref{fig:motivation}, the black-box sampling nature of this computational architecture results in the following uncontrollability manifestations:
\begin{enumerate}
\item \textbf{Black-box probability sampling without world modeling}. Video is projection of $4D$ ($3D + T$) physical world onto $2D$ sensors, but these models directly learn mapping from text towards $2D$ pixels, without constructing an explicit representation of the physical world; they learns statistical correlations, not causal physical rules: "looks right" but "moves wrong", therefore bypassing the $4D$ physical world. They have no notion of objects, properties, or physical laws, \emph{i.e.}, only texture correlations, therefore make physically realistic impossible.

\item \textbf{Extreme dimensional mismatch between sparse instructions and dense outputs}. A text prompt (sparse, low-dimensional) must map to millions of pixels (dense, high-dimensional). This mathematically under-determined mapping forces the model to hallucinate most details, making precise control over specific semantic attributes (\emph{e.g.,} joint angle, camera position and angle) impossible.

\item \textbf{No concept of physical entities, \emph{i.e.}, treating the whole frame as an entangled pixel distribution}. The real world consists of discrete entities (scene, objects, characters) with their own static/dynamic properties. Existing models generate the whole frame as one unified distribution over pixel grids (tensors), \emph{i.e.}, they do not segment or track individual entities. Consequently, there is no way to edit a single instance without affecting the whole image. Geometry, motion, lighting, materials are entangled in black-box latents within the large model.
\end{enumerate}

\section{From End2End Generation to Controller-Renderer Collaboration}
Based on the above analysis, we distill five essential capabilities that a controllable media generation model must possess: 1) \textbf{Explicit $3D$ entity representation}. each object/character must be a separate, trackable $3D$ instance; 2) \textbf{Decoupled control nodes}. Geometry, motion, lighting, and camera parameters must be independently adjustable; 3) \textbf{Editable and visualizable}. Creators must be able to see and directly manipulate the intermediate representation; 4) \textbf{Temporal consistency}. State evolution across frames must be predictable and drift-free; 5) \textbf{Physical plausibility} (optional but beneficial). Basic physical rules (collision, gravity, light transport) can be enforced. These five requirements collectively point to a single conclusion: the ideal model is a $3D + T$ world representation with explicit entities, in other words, a \emph{World Narrative Representation}. Here, "World" captures the explicit $3D$ entity representation, decoupled physical control knobs, and physical plausibility; "Narrative" embodies the temporal consistency, editability, and visualizability that allow creators to craft and refine a story across time.

Rather than replacing existing video foundation models, the proposed new paradigm should work in synergy with them, as shown in Figure~\ref{fig:motivation}. The key insight is to decouple the generation process into two distinct stages: control and rendering, fundamentally shifting away from the end-to-end black-box paradigm.

The \textbf{Controller}, which we term the \emph{World Narrative Model}, is responsible for understanding the user's creative intent and planning the physical evolution of the scene. It produces a structured, low (median) dimensional, and fully editable set of physical parameters: scene graph, asset placements, skeleton motion trajectories, camera path, and lighting configurations. This output is structurally deterministic: given the same input, the same high-level physical parameters (scene layout, entity geometry, subject poses and trajectories, camera patch, lighting setup \emph{etc.}) are produced consistently, providing a reliable blueprint for the renderer.

The \textbf{Renderer}, in contrast, is implemented by existing video foundation models (such as Sora, Kling, Wan, etc.), whose parameters remain frozen. These models act as \emph{neural shaders}: they take the structured physical parameters from the controller and convert them into high-quality pixels. With the structural control provided by the controller, the renderer can generate pixel-level content with greater detail and richness. No retraining of the foundation model is required; control signals are injected via current standard visual reference tokens (such as reference frames), or lightweight trainable adapters which transforms the world narrative representation parameters into model-recognizable latents.

This controller-renderer decoupling can be intuitively understood through an \emph{orchestra analogy}. The conductor (controller) interprets the score and directs the orchestra (renderer) to produce the music. In our framework, the World Narrative Model understands user's intention and plans the "what" and "when" of the physical world, while the video foundation model executes the "how" of pixel rendering, together achieving structure-deterministic, controllable, and high-fidelity video generation that no single model can accomplish alone.

\textbf{Relationship with World Model}. The concept of "world models" originated in embodied AI, where the goal is to enable agents (robots, self-driving cars) to perceive, reason, and act in the physical world. Note that media generation imposes a fundamentally different set of requirements. 
On one hand, embodied AI world models (\emph{e.g.}, JEPA~\cite{Assran_2023_CVPR, assran2025vjepa2}, World Lab's Marble~\cite{worldlabs2025marble}, Meta's WorldGen~\cite{wang2025worldgen}, NVIDIA's Cosmos~\cite{nvidia2025cosmos, nvidia2026cosmos3}) aim to enable agents to make decisions and execute actions in real environments; their outputs are actions, plans, or latent feature representations not designed for human, and they are evaluated by task success rates (\emph{e.g.}, collision avoidance), not by visual quality or editability. In contrast, the proposed world narrative model is designed to empower human creators, \emph{i.e.,} directors, cinematographers, lighting artists, to produce professional media content. Its output consists of visualizable and editable physical parameters that creators can see, inspect, and directly modify, and it must support artistic rendering quality. Evaluation focuses on controllability precision (\emph{e.g.}, how accurately a specified joint angle is realized), editing efficiency (\emph{e.g.}, number of iterations per shot), and user satisfaction.
On the other hand, embodied AI models prioritize strict physical correctness because the simulated world must obey real world physics (gravity, inertia, collision); any deviation could cause a robot to fail or become dangerous, leaving little tolerance for artistic license. Media purposed world model, however, prioritizes controllable artistic expression. While basic physical plausibility (\emph{e.g.}, no object floating in the air) is desirable, creators often need to bend or exaggerate physics for dramatic effect, \emph{i.e.}, a character can jump higher than realistically possible, shadows can be stylized, and lighting can defy natural laws. \textbf{Controllability, not absolute physical fidelity, is the primary metric.}

Compared to world models for embodied AI, the proposed media world model enjoys significant feasibility advantages rooted in its fundamentally different users and application scenarios. Embodied models target robots or autonomous systems, requiring strict closed-loop control; in contrast, media model serves human creators, demanding human-AI collaborative, highly visualizable, and easy-to-use interfaces that seamlessly integrate automatic generation with fine-grained manual adjustment, drastically lowering deployment barriers. Furthermore, embodied models rely on costly, scarce real-world interaction data (\emph{e.g.}, robot manipulation, driving logs), whereas media model can leverage massive, publicly available video archives (films, dramas, short videos) through automatic annotation and intelligent parsing, making data acquisition far more scalable and affordable. Finally, embodied models demand near-perfect physical accuracy (\emph{e.g.}, exact momentum conservation, collision response) to avoid real-world failures; media model only requires visually plausible, artistically controllable physics, \emph{i.e.}, creators can bend gravity, exaggerate motions, or stylize lighting for dramatic effect, namely, it prioritizes artistic intent over physical accuracy. This tolerance for physical approximation greatly simplifies learning and accelerates the transition from lab to production. Consequently, the media world model is substantially more realizable and industrially viable across data, interaction, and physical constraint dimensions. \emph{The above fundamental difference drives all design choices in our World Narrative Model, from its explicit entity representation to its decoupled control nodes and human-in-the-loop interfaces.}

We therefore propose a 
\textbf{World Narrative Model} tailored specifically for media creation, built on an explicit $3D + T + Entity$ representation that directly satisfies the five essential requirements: instance-level 3D scene asset geometry and placements, skeleton motions/trajectories, lighting and camera parameters ($6-DoF$ path, intensity, focal length \emph{etc.}), all spatio-temporally manipulable, with optional physical plausibility (\emph{e.g.}, rigidity, collision, gravity, light propagation). Each object/subject and the associated attribute is a separate, trackable entity, all supporting spatio-temporal manipulation (\emph{e.g.}, changing an object's pose at frame 10, adjusting lighting direction at frame 25). 
Unlike curent video-based world models~\cite{inspatio2026world}, our proposed model is a controller-only architecture: it does not replace existing video foundation models but works with them as frozen renderers. For practical deployability, we avoid retraining large foundation models. Instead, we leverage off-the-shelf CV/CG tools and pre-trained models (such video understanding, $3D$ reconstruction, depth estimation, video object segmentation and tracking \emph{etc.}) and orchestrate them into agent-based workflows. These intelligent agents collaboratively build, manipulate, preview, evaluate and adjust the $3D + T$ world, exchanging structured physical parameters (not pixels). Optionally, lightweight adapters (LoRA~\cite{hu2022lora}, ControlNet~\cite{zhang2023controlnet}) could be trained for conditional injection for video pixel rendering. 

\section{The Proposed Framework: World Narrative Model}
\subsection{Overview}
\begin{figure*}[!t]
\centering
\includegraphics[width=0.95\linewidth]{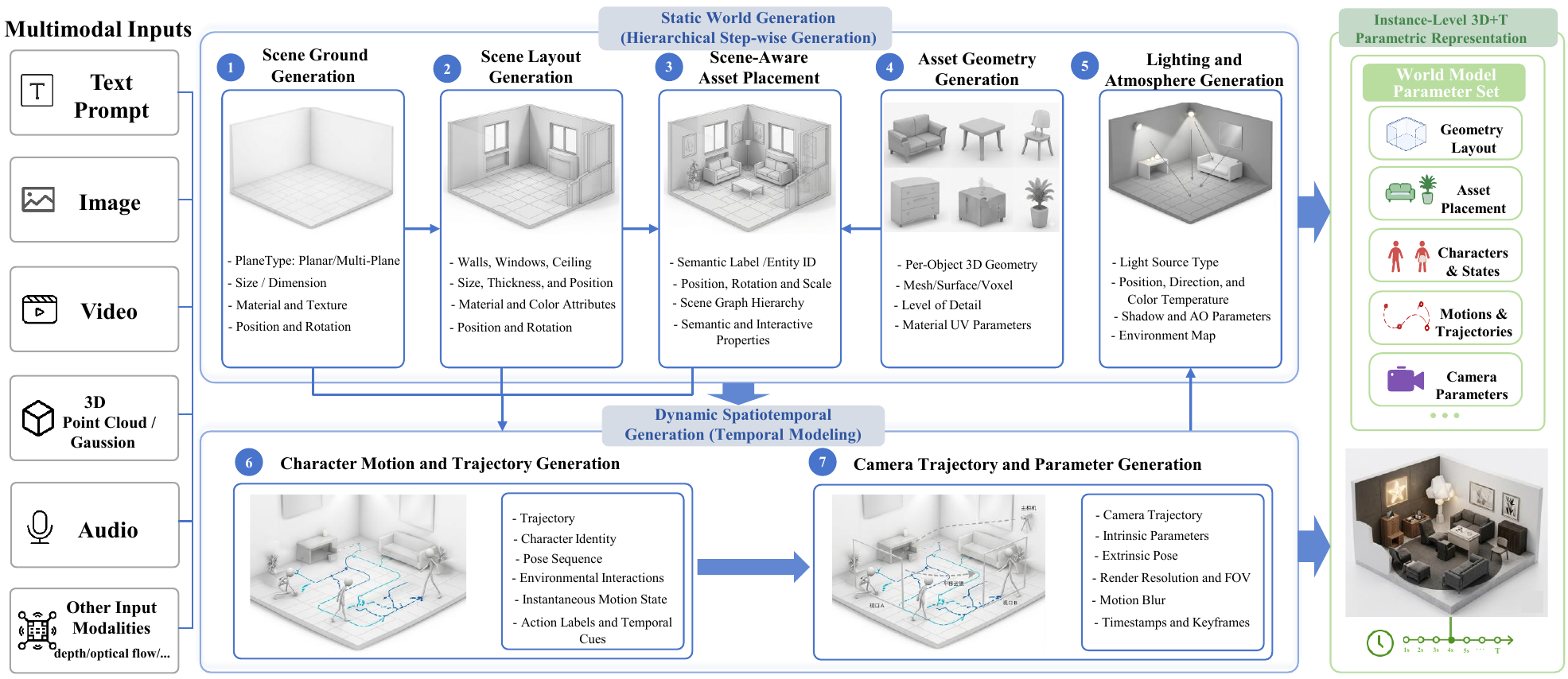}
\caption{An systematic overview of our proposed World Narrative Model. The model is based on a series of collaborating agentic workflows including: scene layout generation, asset generation and placement, motion and trajectory planning, and camera path and lighting parameter setting.}
\label{fig:overview}
\end{figure*}

As depicted in Figure~\ref{fig:overview}, our framework provides a unified pipeline that takes multi-modal inputs (text scripts, reference images/videos, sketches, motion trajectories) and automatically constructs a complete $3D + T$ pre-visualization/storyline scene geometry and layout that matches the text and reference visual constraints, asset geometry (via retrieval or generation balancing efficiency and diversity) with $6-DoF$ placements ensuring physical stability and contextual semantics (collision avoidance, proper support), character motions and trajectories that avoid obstacles, respect terrain, and maintain natural interaction while adhering to the described movement style, camera paths aligned with the narrative cinematography, and lighting configurations consistent with the desired atmosphere. This pre-visualization/storyline serves as a deterministic structure blueprint that can drive any existing video foundation model (frozen) to produce the final video output.

The system operates in two modes: (i) Automatic mode. From inputs to fully generated video without human intervention, suitable for rapid prototyping and storyboarding; (ii) Human-in-the-loop mode. The generated $3D+T$ world is fully editable and inspectable. Professionals (directors, cinematographers, lighting artists) can directly manipulate any physical parameter at any level of detail, from coarse scene layout to fine joint angles, from camera path to light color temperature, using dedicated control panels (3D white-box, skeleton/motion timeline, virtual light console, camera yaw-pitch-trajectory editor) that expose the underlying entity-based representation. Changes are instantly reflected in the pre-visualization and, upon confirmation, fed into the video foundation model for re-rendering.
Thus, the creator works with an interpretable, parameterized/quantified, and fully controllable world representation, \emph{i.e.,}, playing the role as an orchestrator of a physical world simulation, enabling professional-grade, predictable content creation.

\subsection{Scene Layout Agent}
\begin{figure*}[!t]
\centering
\includegraphics[width=0.95\linewidth]{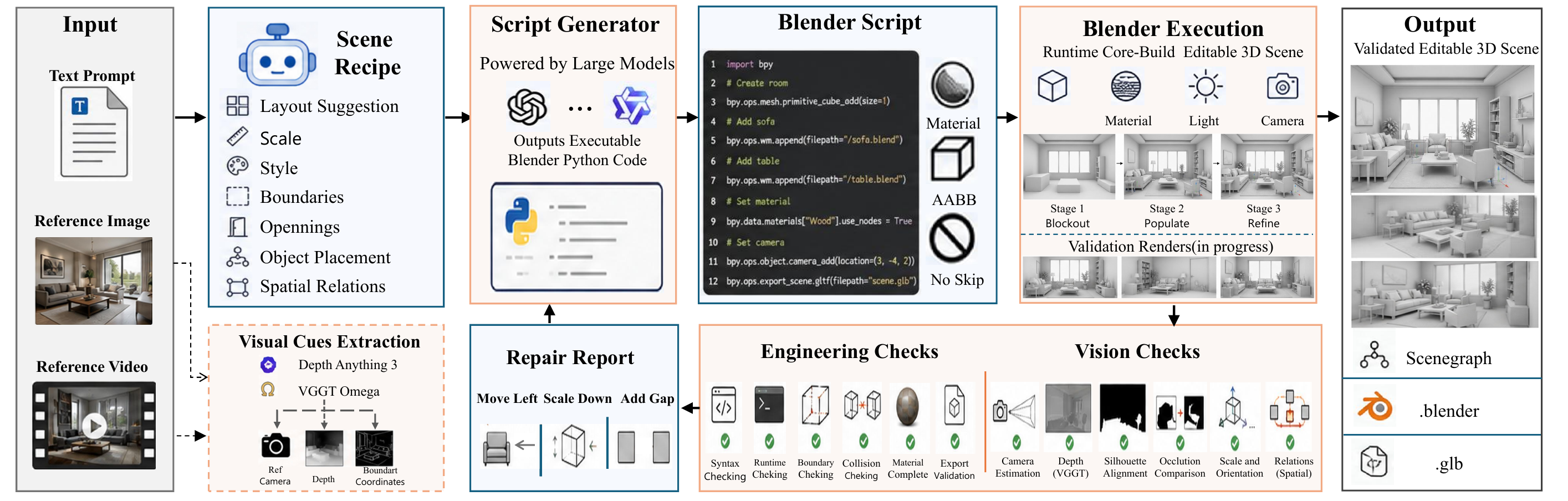}
\caption{Module 1: scene layout generation agentic workflow.}
\label{fig:sceneagent}
\end{figure*}

The first core module of our World Narrative Model is an automatic scene layout generation system that constructs $3D$ environments (including both indoor and outdoor scenes) from multi-modal inputs (text prompts, reference images, or video clips). Unlike conventional single-pass generative methods that produce static meshes or point clouds, our approach adopts a back-to-back multi-agent closed-loop framework that iteratively refines the scene until it satisfies both engineering constraints (collision, boundary, material, exportability) and visual-layout objectives (object placement, spatial relationships, stylistic consistency), as illustrated in Figure~\ref{fig:sceneagent}. The system comprises two collaborating agents: a generator agent and a supervisor agent, working in a closed-loop where text and vision derived constraints are cast as differentiable supervisory signals that drive geometric and layout adjustments in the scene entity generation/parameterization pipeline.

The generator agent takes as input a structured specification derived from the user's description. This specification is not a free-form sentence but a semantically parsed scene recipe, which may include room dimensions, wall segmentation, door/window openings (for indoor settings) or terrain boundaries, vegetation distribution, and landmark placement (for outdoor settings). The agent then produces an executable Blender Python script that constructs the scene procedurally. It handles geometry creation (walls, floors, terrain, furniture primitives, vegetation, \emph{etc.}), material assignment (with Blender node graphs compatible with GLB export), lighting and camera placement (for rendering and check). The generator agent is a large language model (LLM) augmented with Blender API knowledge and constrained by a hard prompt engineering layer that enforces rules: every visible mesh must have a material, objects must register axis-aligned bounding boxes (AABB), and critical elements (main structures or key objects) cannot be skipped.
The system does not treat visual input as a full $3D$ reconstruction target. Instead, it extracts structured layout cues: scene contour, dominant object positions, relative scales, orientation relationships, circulation paths, material palette, lighting atmosphere, and stylistic vocabulary. These cues are merged with the text prompt which defines the functional type ("living room", "garden", "street scene") and object composition, while the vision provides placement and stylistic imitation targets.

The supervisor agent is an orchestration layer that performs static and dynamic cross-modal validation. It runs a series of checks: script syntax, execution log parsing, runtime exceptions, AABB boundary conformance, inter-object collision detection (using Blender's BVH trees), material completeness, and GLB export viability. Beyond engineering constraints, the supervisor also incorporates vision-guided layout supervision: for input images or video sequences, it calls off-the-shelf models (SAM3D~\cite{sam3dteam2025sam3d, yang2026sam3dbody}, VGGT~\cite{Wang_2025_CVPR}, VGGT-Omega~\cite{Wang_2026_CVPR}, Depth Anything 3~\cite{lin2025depthanything3}) to estimate camera parameters, depth maps, and coarse scene geometry. It then renders the current Blender scene from the estimated camera poses and compares the rendered depth, silhouette, object-level occlusion, relative scale, and orientation with the input visual cues. The comparison is not pixel-wise RGB but layout-oriented metrics: depth consistency, contour alignment, relative positional errors, and object-to-object distance discrepancies. These errors are compiled into a structured report and fed back to the generator agent as repair suggestions (\emph{e.g.}, "move the table $0.2m$ left", "reduce the chair size by $5\%$", "add a gap between the sofa and the wall"). The generator then revises the script accordingly. This loop repeats until all constraints are satisfied or a maximum iteration limit is reached.

The main advantages of this module are threefold. First, the back-to-back agent loop transforms scene generation from a "guess-once" black box into an interactive, verifiable, and correctable process that guarantees engineering-level robustness. Second, the integration of vision-guided supervision enables the system to imitate real-world layout patterns from arbitrary images or videos without requiring full $3D$ ground truth. Third, the use of procedural Blender scripts rather than direct mesh generation preserves full editability: the output is a list of parametrized entities with named objects, materials, and camera, which can be further refined by human artists. This module thus provides a practical, deployable solution for automated scene layout that meets the rigorous demands of professional content creation across both indoor and outdoor environments.

\subsection{Asset Generation Agent}
\begin{figure*}[!t]
\centering
\includegraphics[width=0.95\linewidth]{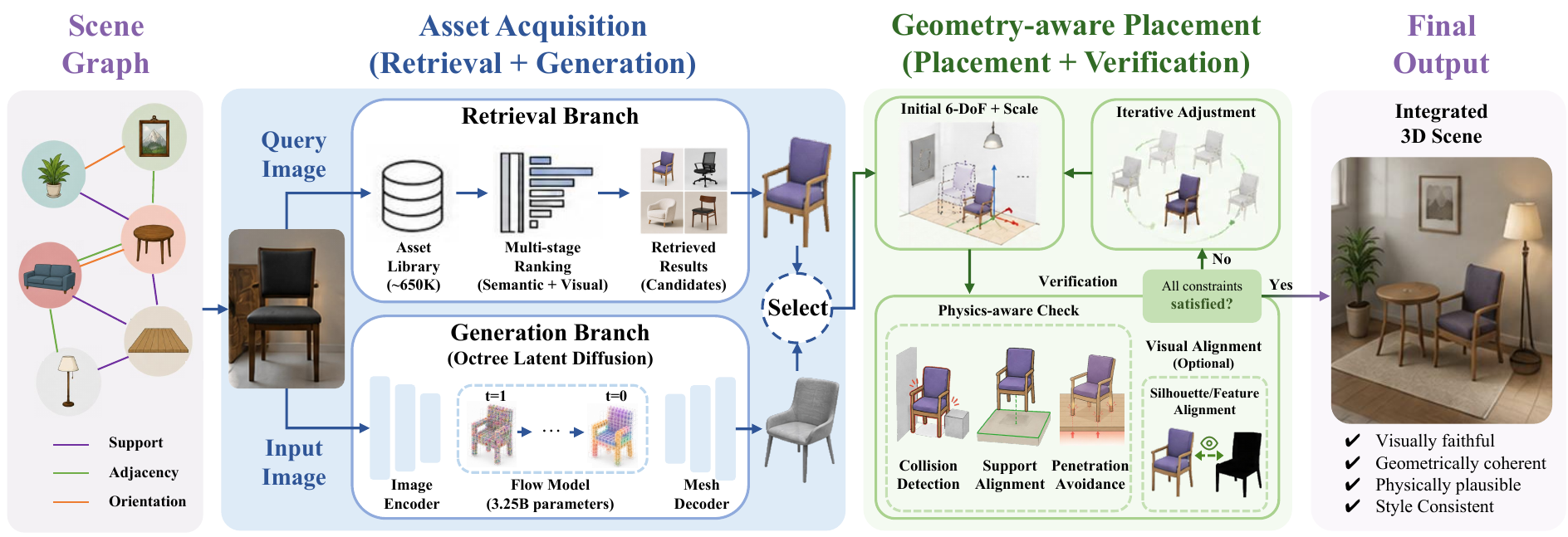}
\caption{Module 2: asset generation and placement agentic workflow.}
\label{fig:assetagent}
\end{figure*}

The second core module of our World Narrative Model handles the creation of $3D$ assets (objects, props, furnishings) and their physically plausible placement into the target scene, as illustrated in Figure~\ref{fig:assetagent}. Unlike conventional methods that generate isolated objects or rely on manual placement, our system operates under a unified scene-graph control that encodes not only the semantics and appearance of each asset but also its spatial relationships (support, adjacency, containment, orientation, scale, and functional connections) with other objects and the scene geometry. The workflow consists of three stages: 1) asset understanding, 2) asset acquisition (retrieval + generation), and 3) geometry-aware placement matching, orchestrated in a closed loop that ensures each asset is both visually faithful to the user's input and geometrically integrated into the scene.

For a given user input (text, reference image, or video clip), a vision-language understanding agent parses the request into a structured scene graph. For text inputs, it extracts object categories, functional attributes, style descriptions, material preferences, size clues, and inter-object relationships. For image or video inputs, the agent first performs semantic discovery to identify observed objects, then uses VGGT-Omega~\cite{Wang_2026_CVPR} to estimate frame-level camera parameters, depth maps, and geometric cues. The masks of each observed object are back-projected into $3D$ space to estimate their spatial position, orientation, scale range, supporting surface, and adjacency relations. The final scene graph is composed of nodes (each representing an object to be generated or matched) and edges (representing support, adjacency, containment, orientation, scale ratio, and functional relationship). Each node carries attributes such as category, material, desired $6‑DoF$ pose, and optional reference imagery.

To efficiently cover both common and long-tail object categories, we adopt a retrieval-augmented generation framework. For each node in the scene graph, the system first queries a high-quality expert asset library (containing approximately $650,000$ pre-normalized $3D$ objects) using a multi-stage ranking algorithm. The retrieval process combines: (i) semantic recall based on category, function, material, and style description via text embedding similarity; (ii) visual similarity matching using image crops from reference input against multi-view rendered thumbnails of library assets; (iii) geometric and scene-constraint re-ranking that matches the bounding-box proportions, support-surface type, and the $6‑DoF$ constraints specified in the scene graph. The final retrieval score is a weighted combination of semantic, visual, and geometric compatibility. If the highest score falls below a threshold, or if the candidate asset fails critical constraints, the system switches to the generation branch.

The generation branch uses a proprietary $3D$ object diffusion model with octree latent representation ($3.25B$ parameters)~\cite{Xiang_2026_CVPR}. By encoding sparse 3D structures in octree latent space, it reduces diffusion sampling cost while preserving geometric detail and topological expressiveness. Compared to state-of-the-art open-source baselines (\emph{i.e.}, Trellis 2.0~\cite{Xiang_2026_CVPR}), this model achieves approximately $40\%$ faster inference at comparable quality, enabling parallel generation of multiple assets in a complex scene. The generated asset is then passed through the same normalization, rendering, and geometric evaluation pipeline as a retrieved asset.

Once an asset is obtained, the placement agent computes an initial $6‑DoF$ transformation from the scene graph constraints, then applies physics-aware checks, collision detection, support alignment, and penetration avoidance. If any constraint is violated, the agent iteratively adjusts the transformation, optionally using a differentiable renderer to match the silhouette to reference imagery. This loop continues until all constraints are satisfied or the iteration limit is reached. The final output is a fully integrated scene where each object is geometrically and physically coherent with the scene.

\subsection{Motion Planning Agent}
\begin{figure*}[!t]
\centering
\includegraphics[width=0.95\linewidth]{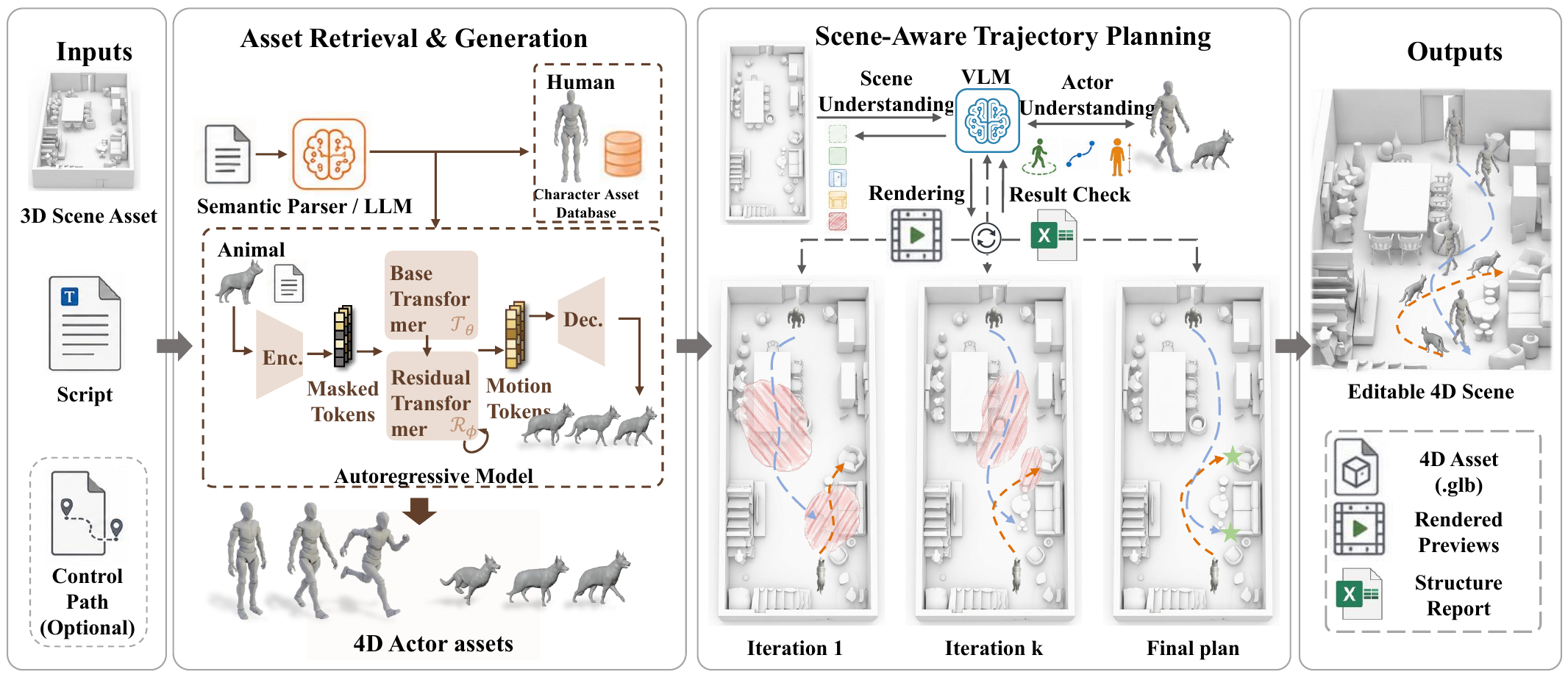}
\caption{Module 3: actor motion and trajectory generation agentic workflow.}
\label{fig:motionagent}
\end{figure*}

The third core module of our World Narrative Model addresses the generation of both skeletal motion and scene-aware trajectory for dynamic entities, including not only human figures but also a wide range of animals (quadrupeds, birds, fish, \emph{etc.} which appear frequently in movie industry), as illustrated in Figure~\ref{fig:motionagent}. Unlike existing methods that focus on a single species or require manual motion re-targeting, our framework is designed for heterogeneous skeletal topologies, namely it accepts as input a target skeleton (joint positions, connectivity, bone lengths) and produces joint-level continuous rotation parameters that drive the asset directly. The motion is generated from text prompts, while the trajectory is automatically planned to respect the physical geometry of the previously constructed 3D scene (\emph{e.g.}, collision avoidance, ground adaptation, natural navigation). The entire system is divided into two synergistic sub-modules: a text-driven motion generator and a vision-language-guided trajectory planner.

The motion generator is built upon a factorized spatio-temporal transformer architecture that separates coarse-grained body motion from fine-grained limb details. First, a residual vector-quantized variational autoencoder (RVQ-VAE)~\cite{lee2022autoregressive} compresses motion sequences into discrete motion tokens at the joint level, using joint padding and masking strategies to handle varying skeleton sizes. This tokenization is trained with a smooth reconstruction loss and a commitment loss applied only on valid tokens. Then, a dual-level transformer predicts the masked motion tokens in a coarse-to-fine manner. The base transformer receives a skeleton-derived prefix (encoding joint type, connectivity, and inter-joint distances) and uses a cosine-scheduled masking ratio to learn the global motion rhythm. The residual transformer takes the cumulative embeddings from previous levels together with level-index embeddings to predict the high-frequency details (\emph{e.g.}, tail wagging, ear twitching). A multi-modal language model (SigLIP2)~\cite{tschannen2025siglip2} provides the text encoder, while a graph neural network encodes the skeletal topology. Together, they condition the transformer to generate motions that are semantically aligned with the text prompt and anatomically plausible for the given species. The model contains $161$ million parameters and was trained on the largest-known animal $4D$ dataset (over $40,000$ motion clips) covering quadrupeds, birds, fish, and other taxa, making it the first to support cross-species motion generation from free-form text.

The trajectory planner takes the initial motion generated by the above network and adapts its placement in the $3D$ scene (\emph{e.g.}, a furnished room or outdoor terrain) to avoid collisions, maintain ground contact, and produce natural movement. We leverage Codex~\cite{openai2025codex} combined with a vision-language model (VLM)~\cite{openai2026gpt55} in a closed-loop "render-evaluate-reprompt" cycle. Given a textual description of the desired movement (\emph{e.g.}, "walk forward then turn backward"), Codex first generates a Blender Python script that places the animated character at a tentative trajectory. The system then performs a dynamic pre-rendering using Blender's API. The rendered frames are fed back to the VLM, which acts as a supervisor: it evaluates the rendered sequence from high-dimensional geometric and semantic perspectives, detecting interpenetration (clipping with scene objects), floating (lack of ground contact), unnatural turning, and violations of scene semantics (\emph{e.g.}, walking through a wall). The VLM then produces a structured error report (\emph{e.g.}, "the left foot intersects the chair leg at frame 45") and automatically rewrites the trajectory parameters. The loop repeats until the scene-aware trajectory is both physically plausible and semantically consistent. This process eliminates the need for manual keyframing and collision clean-up, achieving a fully automatic "text-to-trajectory" pipeline that respects the geometric constraints of the scene.

\subsection{Camera and Lighting Agent}
\begin{figure*}[!t]
\centering
\includegraphics[width=0.95\linewidth]{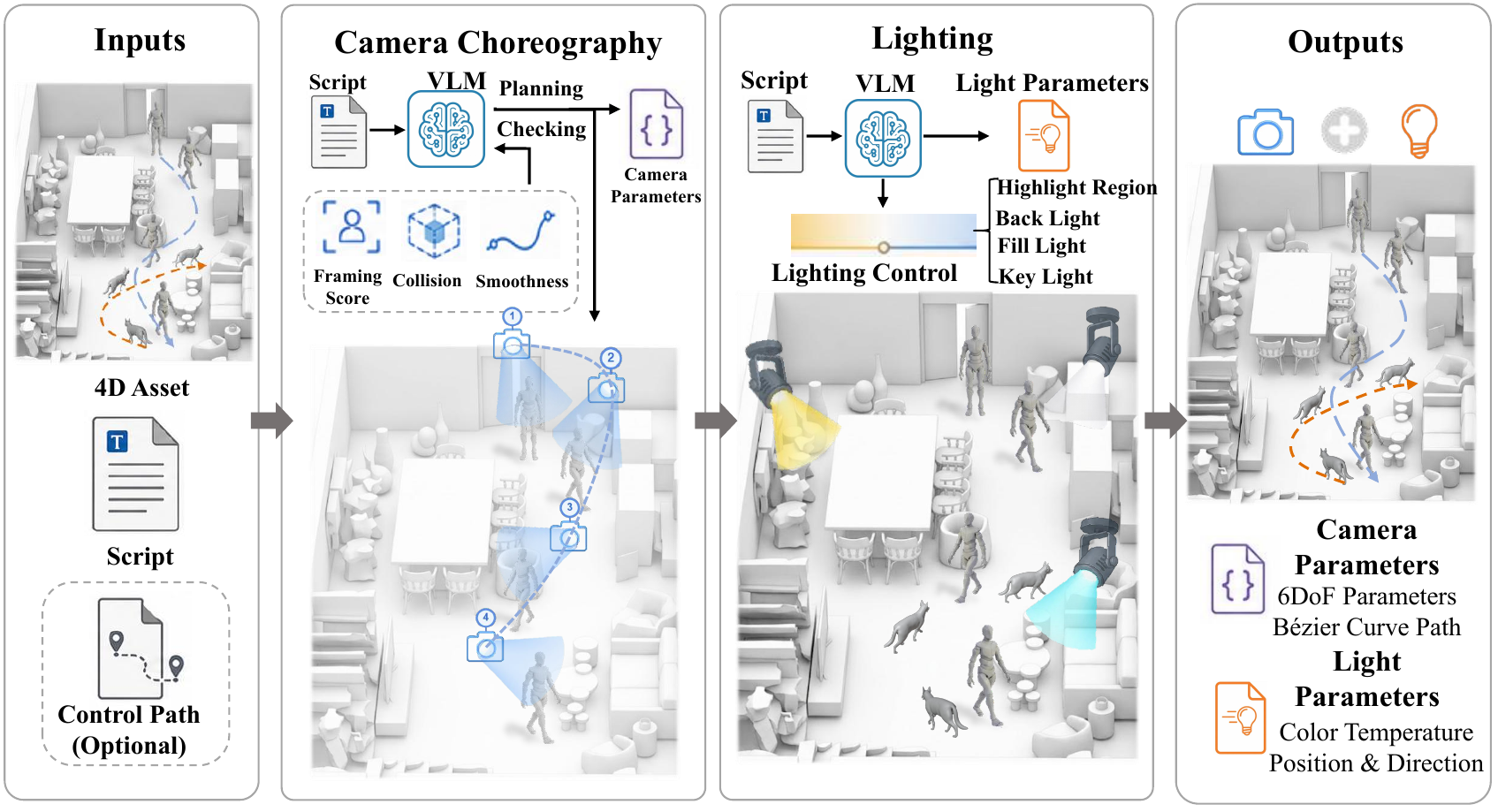}
\caption{Module 4: cinematography and lighting setting agentic workflow.}
\label{fig:cameraagent}
\end{figure*}

As shown in Figure~\ref{fig:cameraagent}, the fourth module automatically generates cinematographic camera trajectories ($6‑DoF$ path, focal length, depth-of-field) and lighting configurations (light positions, types, intensity, color temperature) that are semantically aligned with the textual story description and geometrically compatible with the previously built 3D scene (including object layouts, character motions, and action lines). A camera agent first parses the story line using a large language model to identify key moments, required shot types (\emph{e.g.}, close-up, wide, tracking, over-the-shoulder), and emotional tones (\emph{e.g.}, tense, romantic, dramatic). It then consults a pre-compiled cinematography skill library that maps narrative cues to parametric camera behaviors (\emph{e.g.}, from "dramatic revelation" to slow zoom-in with slight tilt). Using the scene geometry and predicted character/object positions, the agent optimizes the camera trajectory to avoid occlusions, maintain subject framing, and produce smooth motion, finally synchronizing the camera path with the narrative timeline. Similarly, a lighting agent interprets the text description of atmosphere (\emph{e.g.}, "warm sunset", "cold moonlight", "neon noir") and retrieves matching lighting presets (key/fill/back ratios, color temperatures, shadow softness) from a curated lighting skill library. It then places virtual light sources at physically plausible positions in the scene (taking into account object occlusions, surface reflectivity, and desired mood), adjusts their parameters, and optionally refines them via a differentiable renderer to match reference images. The entire process is orchestrated without manual intervention, producing camera and lighting setups that are narratively coherent and scene-aware.

\subsection{A Human-AI Collaborative Video Creation Platform}
\textbf{Overview: Integration of Automation and Direct Manipulation}.
Built atop the World Narrative Engine, the platform provides a friction-free interface where automatic $3D+T$ pre-visualization/storyline and human-directed fine-grained control seamlessly interleave. In auto mode, text or reference materials are turned into a complete world narrative (scene layout, assets, motions, camera, lighting) ready for review. In manual mode, professionals directly manipulate any physical parameter including scene geometry, joint angles, camera path, light color \emph{etc}. at any timestamp; the world state updates instantly, triggering re-rendering. The two modes collaborate in a manner directly aligned with the traditional film production workflow familiar to directors, enabling creators to move effortlessly between high-level automation and precise manual refinement.

\textbf{Interaction Paradigm: Canvas + Node Graph + Narrative Timeline}.
The platform provides three complementary interaction modalities that together support the entire creative pipeline.

\emph{Workflow Canvas}. This is a visual flow graph representing the production pipeline from raw input to final video. Typical nodes include "text to script", "script to world representation", "world to keyframes", and "keyframes to video shot". Within a world representation node, the user can instantiate intelligent control panels, such as the $3D$ white-box for scene/object manipulation, the camera console, the lighting console, and the motion/trajectory console-each embedded as a sub-node in the canvas. These panels expose the full physical parameter space of the World Narrative Model, allowing the creator to directly edit the $3D+T$ pre-visualization at any stage of the workflow.

\emph{Entity Node Graph}. This is a graph of all narrative entities-characters, scenes, props, lights, cameras, and their physical attributes (shape, $6-DoF$ pose, motion style, lighting intensity, color temperature, \emph{etc.}), which compose the underlying representations for the world narrative model. Each attribute is represented as an editable node, and edges capture spatio-temporal relationships among entities (\emph{e.g.}, "character A is standing on platform B", "light L illuminates object C"). Because these edges encode semantic and physical dependencies, modifying one attribute automatically propagates consistent changes to related attributes. For example, moving a character's hand triggers an update of the held object's position and orientation, and changing a light's direction updates the shadows of all affected objects. This ensures both semantic coherence and physical correctness across the entire world state.

\emph{Narrative Timeline}. This is a multi-track timeline that orchestrates the evolution of all entities and their attributes over time. Users can place keyframes for any attribute (\emph{e.g.}, character joint angle, camera position, light color) and smoothly interpolate between them. The timeline supports scrubbing, slicing, copying, and pasting, enabling precise temporal editing. Together with the canvas and node graph, the timeline ensures that the creator can simultaneously control what happens (entity graph), when it happens (timeline), and how it is produced (workflow canvas). All three views are fully synchronized, offering an unprecedented level of flexibility for human-AI collaborative content creation.

\textbf{Director's Control Panels}.
Each control panel is an AI-augmented interface that automates routine tasks while leaving creative decisions to the human. The platform provides four such panels, each dedicated to a specific creative domain, as shown in Figure~\ref{fig:controlpanel}
\begin{figure*}[!t]
\centering
\includegraphics[width=0.98\linewidth]{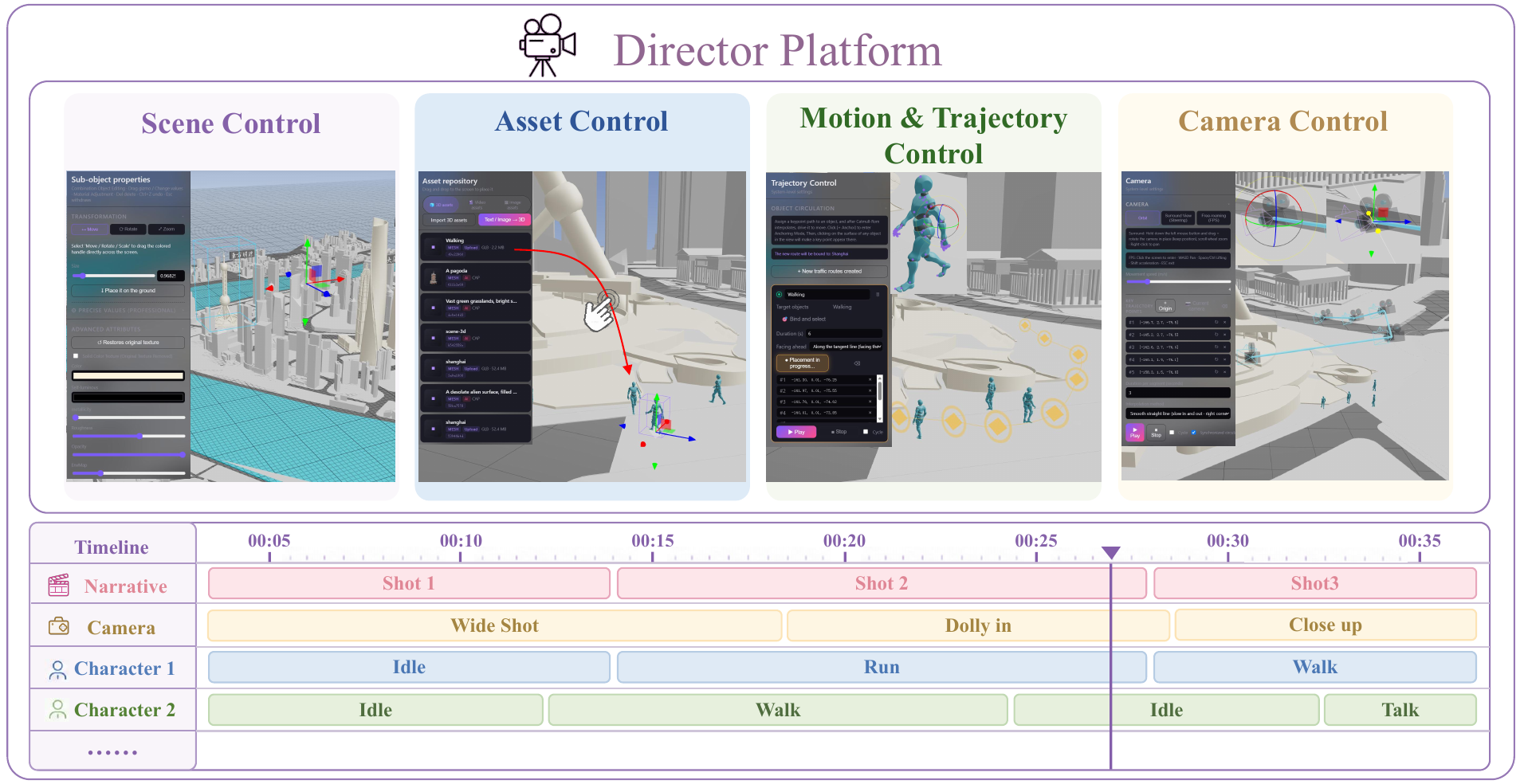}
\caption{A diagram visualization of four director's control panels, including scene, asset, motion and camera manipulations.}
\label{fig:visualize}
\end{figure*}

\emph{Scene and Asset Control Panel}. In auto mode, the panel takes a text description (\emph{e.g.}, "a medieval market square") or a reference video and generates a full $3D$ scene layout with asset placements that respect physical integration (ground alignment, collision avoidance). In manual mode, users can drag objects, change poses, scale, and rotate, assisted by physics-aware feedback (snap to ground, prevent collisions). Assets can be retrieved from a searchable library (including the user's own archive) via semantic queries or generated on-the-fly.

\emph{Character Motion and Trajectory Control Panel}. Auto mode generates skeleton motion (for humans, quadrupeds, birds, fish, \emph{etc.}) and scene-aware trajectories from text (\emph{e.g.}, "a soldier running then jumping") or a reference animation. Manual mode provides a skeleton timeline where joint angles are displayed as curves; editors can drag keyframes or use inverse kinematics to pose characters directly in the 3D canvas, and trajectories can be drawn as B-spline curves on the canvas.

\emph{Cinematography Control Panel}. Auto mode produces a $6-DoF$ camera trajectory with lens parameters (focal length, depth-of-field) from text directives (\emph{e.g.}, "a slow dolly-in with slight tilt") or from a reference video. In manual mode, a camera curve editor based on a Yaw-Pitch-Roll 3D control panel allows users to adjust position and orientation over time by dragging control points, adding keyframes, or drawing the path directly on the canvas as B-spline curves, with real-time preview of depth-of-field effects.

\emph{Lighting and Atmosphere Control Panel}. Auto mode estimates lighting parameters (position, intensity, color, falloff) and can place multiple lights from text (\emph{e.g.}, "warm sunset with backlight") or reference video. In manual mode, a virtual light console shows each light as a draggable 3D gizmo; users adjust intensity, color temperature, and beam angle using sliders or direct canvas dragging, with real-time updates of shadows and highlights.

These four panels, working in concert, give professional creators full control over the physical parameters of the world narrative, while the auto mode provides quick, semantically-aligned starting points for rapid iteration.

\section{Experiment}
\textbf{Overview}.
We design our experiments to address three central questions: 1) Can the World Narrative Model achieve more precise and comprehensive controllability over video generation than existing methods? (2) Does the proposed "physical parameter level control" paradigm genuinely improve the efficiency and experience of professional creators? 3) Does each of the various agentic modules of WNM—scene layout, motion/trajectory, and camera make substantive contributions to the overall controllability? To answer these questions, we conduct a complete evaluation chain that spans from objective metrics to subjective human assessment, with extensive user studies, all within the context of cinematic content creation. The experimental results consistently demonstrate that WNM not only outperforms current state-of-the‑art control paradigms but also provides a practical, deployable solution that meaningfully reduces trial-and-error iterations and empowers creators to achieve deterministic, high-fidelity results.

\subsection{Experimental Setup}
\textbf{Test-bed Construction}.
Our test set is built upon three design principles: diversity of scene types (indoor and outdoor), diversity of motion patterns (single-person and multi-person interactions), and diversity of camera movements (dolly, pan, follow, boom, arc, orbit, \emph{etc.}), ensuring comprehensive evaluation across all controllable dimensions. We source motion data from public action datasets including Mixamo, InterHuman, and InterX, supplement with self-constructed scene graphs and storyboard annotations, and incorporate references from professional film clips. The final test set comprises approximately $3,000$ samples, covering $100$ representative scenes, over $20,000$ action patterns, and 16 distinct camera operation types.

\textbf{Basline Control Methods}. We design three baseline control methods in video generation following a "progressive control difficulty" principle, simulating different user capability levels and the best practices of existing platforms.

1) Text-only baseline (novice mode). This baseline uses only natural language to describe scene, action, and camera requirements, without any reference images or motion priors. It represents the typical usage of mainstream public tools such as native Sora, Kling, and Seedance, namely, "describe-and-sample" control with iterative trial-and-error.

2) Omni-reference baseline (professional mode). This baseline combines multiple modal signals: text prompts, reference images (scene composition and character appearance), reference videos (motion style and camera pacing), and explicit descriptions of camera/trajectory (\emph{e.g.}, "slow dolly-in", "character enters from left"). It represents the strongest control capability supported by current controllable video generation tools (\emph{e.g.}, Seedance Omni-Reference mode), and serves as the upper-bound performance of existing approaches under optimal conditions.

3) WNM (ours). Our method first constructs a coarse $3D+T$ white-box world (including scene proxies, object positions, character stances, motion trajectories, and camera paths) according to the user's control instructions. It then renders a temporal sequence of frames from this white-box world, where each frame encodes explicit physical parameters ($6-DoF$ poses, joint angles, camera viewpoints) as visual proxies. These rendered white-box frames are subsequently used as frame-level control conditions to drive the foundation model for video generation. This is similar to the video2video paradigm adopted by Seedance 2.0, where input frames serve as guidance. However, unlike a real video, our white-box frames contain no photorealistic textures, \emph{i.e.}, they are coarse, explicit, editable, and fully controllable physical proxies. Importantly, we do not train any adapter between the white-box world and the foundation model for this set of experiments; we directly feed the white-box frames as control inputs using the same interface as Seedance 2.0's built-in video2video conditioning. We intentionally avoid adapter training to ensure a fair comparison, isolating the benefit of the physical-parameter-level control paradigm itself, rather than any additional fine-tuning advantage. This setup allows us to directly assess whether the structured, editable white-box representation provides superior controllability. Adapter training remains a future direction for further improvement.

\textbf{Baseline Platforms.}
To ensure fair and comprehensive comparisons, we select the most representative video generation platforms currently available, covering both closed-source commercial models and open-source models:
\begin{enumerate}
\item Seedance 2.0 (including omni-reference mode, supporting multi-modal conditioning);
\item Kling 2.6 (text / reference image / motion brush control);
\item Sora 2 (text-driven, with occasional keyframe conditioning);
\item Wan 2.6 (open-source foundation model, supporting multi-shot and reference video).
\end{enumerate}

Together, these platforms form a baseline spectrum from "pure text" to "multi-reference strong control", covering the strongest controllable video generation capabilities in the current industry.

\subsection{Comparison Protocol and Evaluation Metrics}
Our comparison aims to answer: how does WNM's "physical parameter control" paradigm compare against the best existing control modalities in terms of precision, consistency, and user controllability?

To quantify this, we adopt pairwise comparison with GSB (Good/Same/Bad) scoring, independently judged by professionals with experience in filmmaking or video generation. For each test sample, judges are presented with side-by-side outputs from WNM and a competing method, and asked to assign one of three labels:
\begin{enumerate}
\item \emph{Good}: WNM output is clearly superior in controllability (more precise scene layout, more accurate motion/trajectory, better camera control, \emph{etc.});
\item \emph{Same}: no perceptible difference in controllability;
\item \emph{Bad}: WNM output is inferior  in controllability.
\end{enumerate}
The final result for each pair is determined by majority voting across multiple experts.

We report two types of GSB scores:
1) Overall GSB score: the proportion of comparisons in which WNM is judged as "Good" or "Same" relative to the baseline. This reflects the preference robustness of WNM, \emph{i.e.}, the fraction of cases where WNM is considered at least as good as the competing method.
2) Fine-grained GSB scores: we decompose the evaluation into three control dimensions to reveal per-aspect strengths and weaknesses:
\begin{enumerate}
\item \emph{Scene control}: scene consistency (correct object placement and layout) and temporal consistency (3D structural stability and cross-frame coherence in long videos).
\item \emph{Motion and trajectory control}: motion naturalness and accuracy; trajectory fidelity (whether the character's path precisely follows the specified route); and for multi-person interactions, relative positioning and action synchronization.
\item \emph{Camera control}: trajectory accuracy (smoothness, target tracking stability, and alignment with the intended camera path), as well as matching between camera language and narrative pacing.
\end{enumerate}

In addition, to evaluate the practical value of WNM in real-world production workflows, we introduce efficiency-related auxiliary metrics in the user study. Participants are asked to use WNM to complete specified video creation tasks and rate their experience on a $1–5$ Likert scale across the following dimensions:
\begin{enumerate}
\item \emph{Result quality}: (Q1) The generated video is visually realistic and artifact-free; (Q2) The result faithfully reflects the scene layout and object placements I intended; (Q3) The temporal coherence is satisfactory.
\item \emph{Control accuracy}: (C1) I can precisely specify character actions and movements; (C2) I can precisely control the camera path and shot composition; (C3) I can independently adjust lighting/atmosphere without affecting other aspects; (C4) The system's response to control commands is predictable and deterministic.
\item \emph{Overall usability}: (U1) The learning curve is acceptable; (U2) The tool significantly reduces trial-and-error attempts compared to my usual workflow; (U3) I would use this tool in my professional projects; (U4) Overall satisfaction with the controllability of the system.
\end{enumerate}
Also, the system automatically record objective efficiency metrics: number of generation trials per shot (to measure "gacha" reduction), total creation time (from start until acceptance) per task. 

The above dual-tier design ensures that GSB captures the relative controllability advantage of WNM over existing methods, while the Likert ratings capture absolute user experience and workflow benefits. Both are reported separately and discussed in their respective sections.

\subsection{Quantitative Controllability Results}
\textbf{Overall Comparison Results}.
Table~\ref{tab:gsb} reports the overall GSB scores comparing WNM against each baseline. WNM achieves an overall GSB score of $2.75$ against the text-only baseline (novice mode), indicating that in $73.3\%$ of the comparisons, WNM is judged as superior or equivalent to pure text-driven generation. Against the omni-reference baseline (professional mode), WNM still attains a GSB score of $2.02$, meaning that even when competing against the strongest multi-modal conditioning pipeline currently available, WNM is preferred or tied in $66.9\%$ of the cases. This result is particularly striking because the omni-reference baseline already represents a sophisticated combination of reference images, reference videos, and explicit descriptions, \emph{i.e.}, the best that existing tools can offer. Yet WNM, using only a coarse $3D+T$ white-box world plus a single reference image for appearance, surpasses or matches this strong baseline in the majority of comparisons. These findings collectively validate the superiority of the "physical parameter level control" paradigm over the traditional video generation control paradigm: directly specifying spatial-temporally dense and explicit physical parameters yields more precise and more reliable controllability than composing "sparse" reference signals such as texts and reference images and hoping the model infers the intended constraints.

\subsection{Fine-Grained Control Dimension Results}
Table~\ref{tab:gsb} also breaks down the GSB scores across the three control dimensions. We analyze each dimension in detail below.

\textbf{Scene control}. $WNM$ achieves a GSB score of $3.21$ against the text-only baseline and $2.23$ against the omni-reference baseline in scene control. The advantage is most pronounced in scene consistency (object placement and layout correctness). Text-only generation frequently misplaces objects, \emph{i.e.}, chairs floating above floors, tables intersecting with walls, since the model must infer spatial relations from language alone. The omni-reference baseline improves this by providing reference images, but it still struggles with precise placement when the reference does not exactly match the target layout. In contrast, WNM's explicit $3D$ scene graph encodes object positions, orientations, and scales in absolute coordinates, ensuring that every object is placed exactly where the creator specified. For temporal consistency, WNM also outperforms both baselines: because the white-box world maintains a persistent $3D$ state across frames, objects do not drift or deform as the video progresses. The omni-reference baseline, lacking such a persistent state, often exhibits gradual shape distortion or position drift over longer sequences.

\textbf{Motion and trajectory control}. WNM achieves GSB scores of $2.64$ and $2.02$ against the text-only and omni-reference baselines, respectively. For action control, the gap is largest in multi-person interactive scenarios (\emph{e.g.}, boxing, dancing, or collaborative tasks). Text-only generation often fails to maintain the correct relative positions and synchronization between interacting characters; the omni-reference baseline improves this by using reference videos, but the alignment remains approximate because the reference motion is not explicitly parameterized. WNM, by specifying each character's joint angles and relative positions as explicit functions of time, ensures precise synchronization and spatial separation, preventing the common failure cases of characters passing through each other or losing eye contact. For trajectory control, WNM’s advantage is similarly evident: when a character must navigate a narrow path with multiple turns, the explicit trajectory waypoints guarantee that the character follows the intended route, whereas the baselines (even with reference videos) often deviate or clip through obstacles.

\textbf{Camera control}. WNM achieves GSB scores of $3.33$ and $2.35$ against the text-only and omni-reference baselines in camera control. The advantage is most pronounced for complex, dynamic camera movements such as orbiting shots, drone flyovers, or sustained tracking shots. Text-only descriptions like "fly over the city from east to west" leave too much ambiguity for the model to infer the exact path, altitude, and orientation changes. The omni-reference baseline can copy a camera path from a reference video, but it cannot adapt the path to different scene layouts or storytelling needs, \emph{i.e.}, it is essentially a "copy-paste" operation. WNM, in contrast, allows the creator to explicitly define camera waypoints and interpolation curves, ensuring that the camera follows the exact intended trajectory while adapting smoothly to the scene geometry and narrative pacing.

\subsection{Qualitative Results}
To qualitatively illustrate WNM’s control advantages, we examine six representative scenarios that collectively cover the full spectrum of controllability dimensions. The visualization results are shown in Figure~\ref{fig:visualize}.
\begin{figure*}[!t]
\centering
\includegraphics[width=0.98\linewidth]{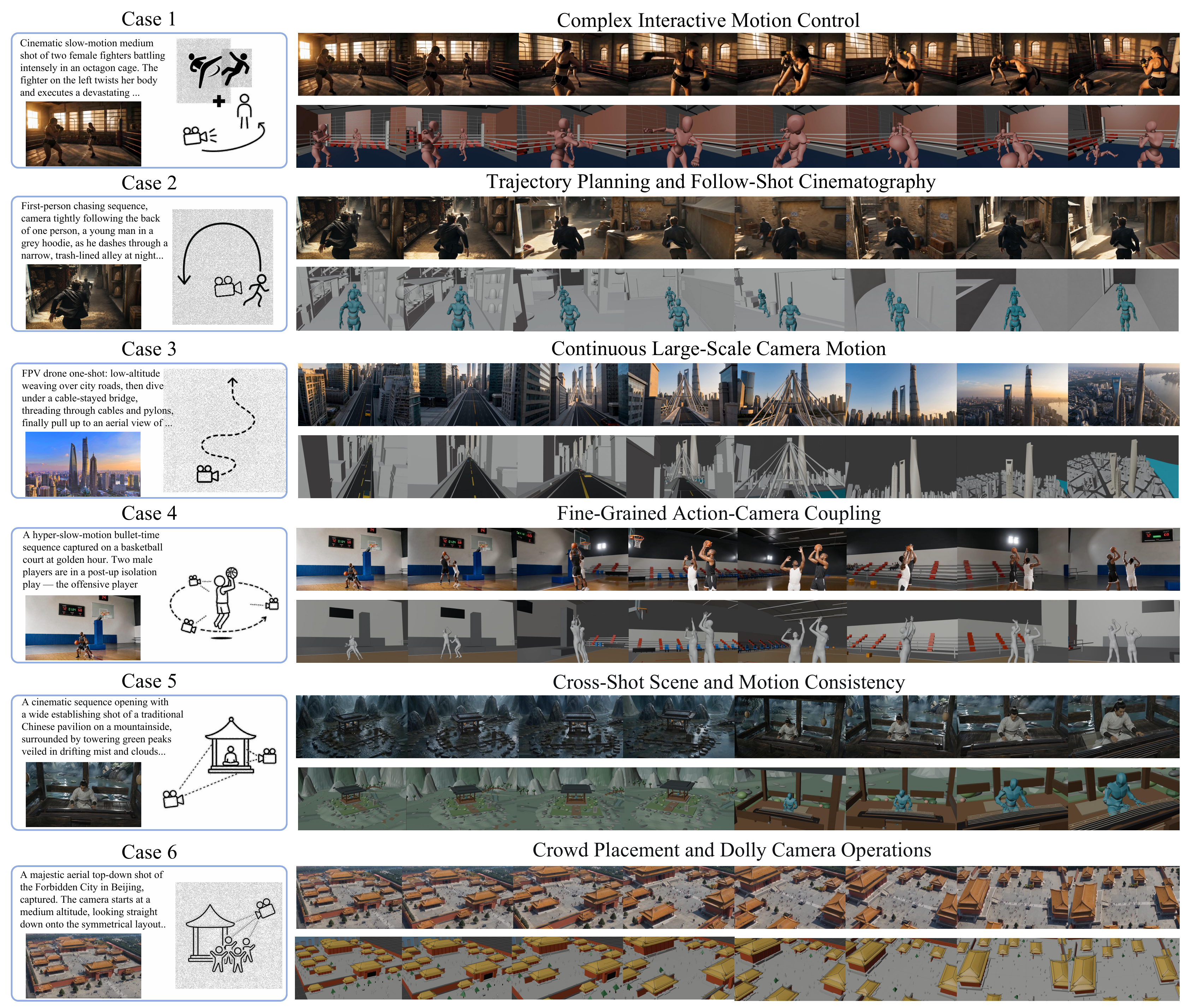}
\caption{Visualization of six representative video generation results by using WNM as control. The upper rows illustrate the rendered video frames by Seedance 2.0, and the lower rows indicate the corresponding (temporally aligned) WNM representations ($3D + T$).}
\label{fig:visualize}
\end{figure*}

Indoor boxing ring fight. In this scenario, two boxers engage in close-range combat with rapid punches, dodges, and clinches. The challenge lies in maintaining precise skeletal articulation for each character while ensuring their limbs do not inter-penetrate. WNM accurately positions each joint at every frame, keeping the boxers' gloves aligned with the target trajectories and their bodies at the correct distance. The baseline methods can produce visually plausible boxing scenes, but they fail to follow the director's specified movement trajectories and detailed punching actions: the characters' footwork deviates from the planned paths, and the timing and targeting of punches cannot be precisely controlled according to the script. This is because they lack explicit joint-level trajectory control and scene-aware spatial reasoning. WNM, by contrast, enforces the exact joint trajectories, spatial positions, and interaction timing as specified in the white-box world, ensuring that every movement and positioning decision is under the creator's direct control.

Narrow street chase. This scene features a character sprinting through a winding alley with multiple sharp turns, captured by a continuous follow-shot camera. The primary difficulty is ensuring that the character's path precisely follows the street layout while the camera maintains a stable following view. WNM’s trajectory waypoints are automatically placed along the street centerline or exactly specified by the director, and the camera path is synchronized to track the character's position in a classic follow-shot manner, resulting in a smooth, coherent, cinematic chase sequence following the planned geometry. The omni-reference baseline, even with a reference video of a similar chase, cannot precisely adapt the trajectory to the specific street geometry; the resulting motion path tends to be stiff and unnatural: the character's turns are not smoothly aligned with the street bends, and the camera angles deviate from the intended framing, failing to capture the continuous tracking effect that the director envisioned. 

Urban drone flyover. This scenario requires a drone camera to fly over a cityscape with multiple altitude changes, direction shifts, and speed variations. The generated video must maintain spatial coherence (buildings should not wobble or shear) and cinematic quality (smooth motion, appropriate framing). WNM's explicit camera waypoints and spline interpolation produce a smooth, stable flyover that respects the city's 3D layout. Baseline methods struggle with this task: text-only generation lacks the spatial precision to maintain consistent building geometry; the omni-reference baseline can copy a reference drone shot but cannot adapt the path to a different city layout, often resulting in misaligned landmarks and unnatural camera motion.

Basketball slam dunk with bullet-time $360°$ rotation. This scene features a player performing a slam dunk while the camera executes a $360°$ orbit around the player, freezing the action at the peak of the dunk. The key challenge is synchronizing the camera motion with the player's pose, \emph{i.e.}, the bullet-time effect requires the camera to rotate smoothly while the player's articulation (jump height, arm extension, ball position) is precisely controlled. WNM specifies both the camera orbit parameters and the player's joint angles as explicit functions of time, ensuring perfect synchronization: the camera completes its $360°$ orbit exactly as the player reaches the apex of the jump. Baseline methods produce visible de-synchronization: either the camera rotation is too fast or too slow relative to the action, or the player's pose drifts during the rotation.

Ancient pavilion with qin music. This scene begins with a wide shot of a tranquil pavilion on a hill, with a robed figure playing a qin (zither), and gradually zooms in to a close-up of the figure's hands and the instrument. The challenge is two-fold: the zoom-in must smoothly reveal fine details (the qin's wood grain, the player's sleeve patterns, the surrounding vegetation) without losing compositional coherence; and the pace of the zoom must match the narrative rhythm. WNM encodes the camera zoom curve and the target framing as explicit parameters, allowing the system to smoothly transit from wide shot to close-up while maintaining spatial consistency. The omni-reference baseline, even with a reference video of a similar zoom, often produces abrupt framing shifts or loses detail due to the model's inability to precisely control the camera parameters.

Forbidden City grand ceremony. This complex scene involves hundreds of participants arranged in precise formations on the palace square, with a camera that alternately pulls back for a wide establishing shot and dollies in for close-ups of key figures. The primary challenges are multi-fold: ensuring that each participant is positioned correctly relative to the architecture and to one another; coordinating individual trajectories (\emph{e.g.}, officials walking in procession); and synchronizing complex camera movements with the crowd dynamics. WNM's scene graph explicitly encodes the position, orientation, and trajectory of every participant, while the camera path is specified as a sequence of waypoints with programmable transitions. The result is a coherent, large-scale scene where the crowd behaves exactly as scripted and the camera moves seamlessly between wide and tight shots. Baseline methods, which lack explicit entity-level control, produce muddled crowds with overlapping figures, missing or duplicated participants, and camera motions that do not align with the scene composition.

In every above representative scenarios, the qualitative results confirm the quantitative GSB findings: WNM's explicit, editable, white-box world representation provides a level of control precision, consistency, and flexibility that current baselines, even the strongest omni-reference configurations, cannot achieve. The primary reason is that reference-based approaches operate through indirect approximation, whereas WNM operates through direct specification. This fundamental difference is most apparent in scenarios that require precise spatial relationships (narrow streets, large crowds), coordinated multi-entity dynamics (boxing, bullet-time), or complex camera-scene coupling (drone flyovers, zoom-ins). We emphasize that WNM achieves these results without any adapter training, using only the white-box frames as control inputs highlighting the inherent power of the world narrative paradigm itself.

\subsection{User Study Results}
Table III reports the user study results comparing WNM against the baseline platforms (text-only for novice users, omni-reference for professional users). For each participant group, we record the number of generation attempts per shot, total creation time (minutes), and subjective ratings (1-5 Likert scale) across three dimensions.

\begin{table*}[t]
    \centering
    \footnotesize
    \setlength{\tabcolsep}{2.6pt}
    \renewcommand{\arraystretch}{1.12}
        \caption{GSB score comparsion between WNM and different baselines. We report both overall GSB scores and per evaluation item GSB scores.}
        \begin{tabular*}{\textwidth}{@{\extracolsep{\fill}}L{4.55cm}C{1.45cm}C{1.55cm}C{1.45cm}C{1.45cm}C{1.45cm}C{1.70cm}@{}}
            \toprule
            \multirow{2}{*}{\textbf{Comparative Setting}} &
            \multirow{2}{*}{\textbf{\shortstack{Overall \\GSB Score}}} &
            \multicolumn{2}{c}{\textbf{Scene Control}} &
            \multicolumn{2}{c}{\textbf{Motion and Trajectory Control}} &
            \textbf{Camera Control} \\
            \cmidrule(lr){3-4}\cmidrule(lr){5-6}\cmidrule(l){7-7}
            & & \textbf{\shortstack{Scene\\Consistency}} &
            \textbf{\shortstack{Temporal\\Consistency}} &
            \textbf{\shortstack{Motion\\Control}} &
            \textbf{\shortstack{Trajectory\\Control}} &
            \textbf{\shortstack{Camera Path\\Precision}} \\
            \midrule
            WNM + Novice \emph{vs.} Novice Baseline & 2.75 & 3.21 & 2.59 & 2.64 & 2.97 & 3.33 \\
            WNM + Professional \emph{vs.} Professional Baseline & 2.02 & 2.23 & 1.93 & 2.02 & 2.12 & 2.35 \\
            WNM + Novice \emph{vs.} Professional Baseline & 1.72 & 2.02 & 1.68 & 1.60 & 1.76 & 1.86 \\
            \bottomrule
        \end{tabular*}
        \label{tab:gsb}
\end{table*}

The user study results show a consistent and substantial reduction in both the number of generation attempts and the total creation time when using WNM, across both novice and professional user groups. Baseline methods, whether text-only or omni-reference, require iterative guess-and-check cycles because they rely on indirect approximation, \emph{i.e.}, the user describes the intent and the model samples from a probability distribution, with no mechanism for precise, parameter-level correction. WNM breaks this loop by enabling direct specification and editing of the underlying $3D+T$ physical parameters; each adjustment targets a specific attribute deterministically, eliminating the need for repeated regeneration. This translates directly to the measured improvements: novices reduce attempts from $18.6$ to $5.1$ ($-72.6\%$) and professionals from $9.4$ to $3.2$ ($-66.0\%$), with corresponding time savings of $-59.8\%$ and $-48.4\%$. Subjective ratings also improve across all dimensions, with the largest gains in control accuracy, \emph{i.e.}, the core capability that differentiates direct specification from indirect prompt probe.

\begin{table*}[t]
    \label{tab:user}
    \centering
    \footnotesize
    \setlength{\tabcolsep}{3.2pt}
    \renewcommand{\arraystretch}{1.18}
    \begin{threeparttable}
        \caption{User efficiency and satisfaction comparison: using \emph{vs.} not using WNM}
        \begin{tabularx}{\textwidth}{@{}L{1.9cm}L{2.0cm}Y Y Y Y Y@{}}
            \toprule
            \textbf{User Group} &
            \textbf{Setting} &
            \textbf{\shortstack{Trials pre Shot\\($\downarrow$)}} &
            \textbf{\shortstack{Creation Time\\min ($\downarrow$)}} &
            \textbf{\shortstack{Result Quality\\($\uparrow$)}} &
            \textbf{\shortstack{Control Accuracy\\($\uparrow$)}} &
            \textbf{\shortstack{Overall Usability\\($\uparrow$)}} \\
            \midrule
            \multirow{2}{*}{Novice}
            & Without WNM & 18.6 & 74.2 & 3.2 & 2.4 & 2.8 \\
            & With WNM & \textbf{5.1} & \textbf{29.8} & \textbf{4.2} & \textbf{4.3} & \textbf{4.1} \\
            \midrule
            \multirow{2}{*}{Professional}
            & Without WNM & 9.4 & 46.7 & 3.8 & 3.4 & 3.5 \\
            & With WNM & \textbf{3.2} & \textbf{24.1} & \textbf{4.4} & \textbf{4.6} & \textbf{4.3} \\
            \bottomrule
        \end{tabularx}
    \end{threeparttable}
\end{table*}

Meanwhile, professionals show less absolute improvement in time despite similar attempt reduction. Professional users already possess better prompt engineering skills and a stronger intuition for what the model can infer from reference signals. Their baseline attempt count ($9.4$) is already lower than novices ($18.6$), so the absolute reduction in attempts ($6.2$ \emph{vs.} $13.5$) is smaller. Moreover, professionals spend a significant portion of their time on high-level creative decisions (shot composition, narrative pacing, stylistic choices) that remain equally time-consuming regardless of the tool. WNM accelerates the execution of their intent but does not eliminate the creative thinking phase. Nevertheless, the $48.4\%$ reduction in total creation time (from $46.7$ to $24.1$ minutes) is still substantial and practically meaningful for professional production workflows.

Subjective ratings confirm the usability advantage. The largest subjective improvement for both novice and professional users is in control accuracy ($+1.9$ and $+1.2$ points, respectively). This aligns with the quantitative attempt reduction: users feel that they can more precisely express their intentions because they are working directly with physical parameters rather than natural language approximations. The overall usability rating also increases substantially, particularly for novices ($+1.3$ points), suggesting that the direct manipulation paradigm is more approachable than the "prompt engineering" paradigm for users without extensive AI experience.

\subsection{Future Experimental Plan}
While the current user study provides strong initial evidence for WNM's effectiveness, the sample size and task diversity are limited. To rigorously validate WNM for industrial deployment, we plan a large-scale user experiment involving at least 10 professional content creation studios (specializing in short dramas, advertising, and cinematic production) and student amateur groups from multiple universities. This experiment will focus on the following dimensions that are difficult to adequately assess in a small-scale lab setting.

\textbf{Long-term human-AI collaboration effects}. We will track the learning curves and efficiency trajectories of users across multiple consecutive projects (each lasting 1-2 weeks of production work). For each user group, \emph{i.e.}, novices, intermediate, and professionals, we will record per-project metrics (attempt count, creation time, subjective satisfaction) over a 3-month period. This will reveal whether the productivity gains saturate quickly or continue to improve as users develop a deeper understanding of the white-box editing paradigm. It will also show whether the tool's learning curve is steep or gradual for different skill levels, informing the design of onboarding and training materials.

\textbf{Cross-scenario generalization}. We will extend the test set to cover a wider range of genre settings-science fiction, historical drama, disaster, fantasy, and period pieces as well as significantly longer narrative segments ($\leq5$ minutes) that require sustained continuity across multiple scene changes. This will assess WNM's robustness to out-of-distribution inputs (\emph{e.g.}, futuristic cityscapes, ancient temples, post-apocalyptic ruins) and its ability to maintain temporal coherence over extended sequences. We are particularly interested in whether the explicit white-box representation inherently generalizes better than reference-based methods, which are more likely to overfit to the styles and layouts present in their training data.

\textbf{Industrial pipeline compatibility}. A critical factor for real-world adoption is whether WNM's outputs, \emph{i.e.}, structured scene graphs and physical parameter sequences, can be seamlessly imported into existing CG pipelines (\emph{e.g.}, Maya for animation, Unreal Engine for real-time rendering, Nuke for compositing). In this experiment, we will task professional artists with taking WNM's outputs and performing typical post-generation operations: relighting a scene, replacing a character model, adjusting a camera path, or adding visual effects. We will measure the additional time required for these operations compared to starting from scratch, and we will document the specific data format requirements needed to ensure smooth interoperability. This experiment will directly inform the design of export APIs and file formats for WNM.

\textbf{Foundation video model adapter training.} We also plan to develop and train a dedicated adapter interface that maps structured world narrative parameters (scene graph, joint angles, camera waypoints, lighting) into the internal representation space of foundation models, enabling even tighter integration. We will leverage an automatic annotation pipeline to generate large-scale paired data of "world narrative representation and video clip", then apply parameter-efficient fine-tuning (LoRA/Adapter) with physical consistency losses to train the adapter while keeping the foundation model frozen. This will further improve control fidelity, particularly for fine-grained attributes, and will extend our approach to multiple foundation models (\emph{e.g.}, Kling, Wan) under a unified control protocol, paving the way for broader industrial deployment.

\section{Future Directions}
\subsection{Current Limitations and Planned Improvements}
While our framework establishes a preliminary foundation for controllable video generation via world narrative models, it also reveals several limitations that point toward promising future research directions.

First, the precision and robustness of agent-based $4D$ content generation remain insufficient for professional production. The current agent pipeline for scene layout, motion/trajectory planning, and cinematography/lighting produces results that are semantically plausible but often lack the fine-grained accuracy demanded by real-world filmmaking. Another limitation of the current system is that motion/trajectory generation and object placement are handled separately, without explicit modeling of human-object interaction. For example, a character picking up a cup or pushing a chair requires coordinated generation of both the character's articulation and the object's $6‑DoF$ trajectory, which is not supported in the current version. The root cause lies in two factors: (i) the agent relies primarily on LLM-based~\cite{openai2026gpt55} implicit reasoning for spatial understanding, which is inherently coarse and lacks explicit geometric computation; (ii) the agent training (fine-tuning) data, though extensive, does not yet cover the full diversity of film production scenarios, including complex interactions between characters and objects, dynamic scene changes, and subtle emotional cues conveyed through camera and lighting.

To address these, we plan the following improvement strategies. First, we will augment the agent with explicit spatial computation capabilities by integrating a rich set of geometry APIs and skill libraries that operate through direct mathematical reasoning rather than implicit LLM inference. For example, camera placement can be optimized via explicit visibility and occlusion solvers; character trajectories can be refined using physics-based path planning that respects terrain geometry and dynamic obstacles. Second, we will significantly expand the training data in multiple dimensions: (a) collect more professional film assets with higher geometric detail (including close-up facial expressions, hand gestures, and detailed costume deformations); (b) annotate more fine-grained action and motion categories especially those involve human-object manipulations; (c) curate and encode a large library of cinematography "skills" from professional films shot composition rules, lighting ratios, color palettes for different genres, which can be retrieved and composed by the agent.

\subsection{Standardization and Benchmarking Development}
Beyond our own technical improvements, we believe that the emerging paradigm of world model for media generation requires urgent community-wide standardization efforts in three interconnected areas. These standards will not only accelerate research but also enable interoperability across different systems, lower the barrier for adoption, and ensure that the technology can be trusted and certified for professional use.

First, standardization of \textbf{world narrative representation}. We envision a hierarchical, level-based specification: from $L0$ (coarse scene graph with object categories and bounding volumes) to $L4$ (detailed geometry, physics properties, material definitions, and temporal behaviors). Each level defines a clear set of required and optional attributes (shape, material, motion, geometry) ranging from coarse parameterization to fine-grained representation, from partial physical functionality towards full functionality. Such a hierarchical standard enables scalable deployment across diverse use cases, \emph{i.e.}, coarse levels for rapid prototyping, finer levels for high-fidelity production, striking an optimal balance between precision and efficiency. This flexible backbone lowers the barrier for third-party developers and fosters a thriving ecosystem of compatible tools, asset libraries, and evaluation suites. By providing concrete data schemas and open-source reference implementations, we aim to unlock broader industrial applications and accelerate the adoption of world-narrative-based content creation across the media industry.

Second, standardization of \textbf{human-machine control interfaces} for video generation. We distinguish two aspects: (a) the algorithmic control interface, which defines what modalities of control signals the generation system should accept (text, reference images/videos, 3D sketches, motion trajectories, direct parameter specifications) and in what format (\emph{e.g.}, structured JSON for camera parameters, BVH for motion, OpenEXR for lighting); and (b) the user-side control interface, which defines what parameters and interaction elements should be exposed to different creative roles. For a director, the interface should expose story-level controls (pace, emotion curve, shot composition); for a cinematographer, it should expose camera trajectory parameters, lens settings, and depth-of-field controls; for a lighting artist, it should expose light types, positions, intensities, color temperatures, and shadow parameters; for a motion director, it should expose joint angles, trajectory curves, and interaction constraints. We propose standardizing both the data formats and the semantic definitions of these parameters that can be understood by different generation systems, so that control panels, agents, and tool chains from different vendors can interoperate seamlessly.

Third, standardization of \textbf{controllability quality evaluation metrics}. Evaluating controllability is inherently more complex than evaluating visual quality or semantic alignment. We propose a multi-level controllability benchmark that defines: (i) controllability levels (from $L0$: binary presence/absence of control, to $L4$: full, fine-grained, decoupled control over all physical parameters); (ii) a set of standardized test tasks that cover different control dimensions, \emph{i.e.}, geometry control (specify object position, size, orientation), motion control (specify joint angles, trajectories), camera control (specify $6‑DoF$ path, focal length), lighting control (specify light positions, colors, intensities), and combined control (simultaneously modify multiple attributes); (iii) quantitative metrics for each dimension, such as mean joint angle error, camera path deviation, lighting consistency score, and decoupling score (measuring how much editing one attribute affects others); and (iv) a human-in-the-loop evaluation protocol, where professional creators are asked to perform specific editing tasks and their effort (number of edits, time, satisfaction) is recorded. We encourage the community to contribute to an open controllability benchmark that is continuously updated, allowing direct comparison of different approaches and driving progress toward truly controllable content creation. 

\section{Conclusion}
We have introduced a fundamentally new approach to video generation control, establishing a World Narrative Model framework that replaces indirect prompt-based sampling with direct physical parameter specification. Built upon this framework, we have developed a human-AI collaborative creation platform that enables deterministic, editable, and interactive content production. Extensive experiments demonstrate that this new paradigm significantly improves controllability and creator efficiency across diverse production scenarios. We envision this work as a starting point for a broader ecosystem, and we encourage the community to contribute new modules, evaluation benchmarks, and standardization efforts to advance the field toward industrial-scale deployment.
%
%
%

\section*{Acknowledgments}
We thank datacanvas.com for providing the computational resources that supported this project throughout its entire lifecycle, \emph{i.e.}, from model training and system deployment to external testing and user studies. Their generous support was instrumental in bringing the World Narrative Model framework from research prototype to a publicly accessible platform.

\bibliographystyle{IEEEtran}
\bibliography{wnmbib}

@article{liu2024sora,
      title={Sora: A Review on Background, Technology, Limitations, and Opportunities of Large Vision Models},
      author={Liu, Yixin and Zhang, Kai and Li, Yuan and Yan, Zhiling and Gao, Chujie and Chen, Ruoxi and Yuan, Zhengqing and Huang, Yue and Sun, Hanchi and Gao, Jianfeng and He, Lifang and Sun, Lichao},
      year={2024},
      eprint={2402.17177},
      archivePrefix={arXiv},
      primaryClass={cs.CV},
      journal={arXiv preprint arXiv:2402.17177},
      doi={10.48550/arXiv.2402.17177},
      url={https://arxiv.org/abs/2402.17177}
}

@misc{teamseedance2025seedance15pro,
  title         = {{Seedance 1.5 pro}: A Native Audio-Visual Joint Generation Foundation Model},
  author        = {{Team Seedance} and others},
  year          = {2025},
  eprint        = {2512.13507},
  archivePrefix = {arXiv},
  primaryClass  = {cs.CV},
  doi           = {10.48550/arXiv.2512.13507},
  url           = {https://arxiv.org/abs/2512.13507},
  note          = {Seedance 1.5 pro Technical Report}
}

@misc{teamseedance2026seedance2,
  title         = {{Seedance 2.0}: Advancing Video Generation for World Complexity},
  author        = {{Team Seedance} and others},
  year          = {2026},
  month         = apr,
  eprint        = {2604.14148},
  archivePrefix = {arXiv},
  primaryClass  = {cs.CV},
  doi           = {10.48550/arXiv.2604.14148},
  url           = {https://arxiv.org/abs/2604.14148},
  note          = {Seedance 2.0 Model Card}
}

@article{gao2025seedance,
  title         = {{Seedance 1.0}: Exploring the Boundaries of Video Generation Models},
  author        = {Gao, Yu
                   and Guo, Haoyuan
                   and Hoang, Tuyen
                   and Huang, Weilin
                   and Jiang, Lu
                   and Kong, Fangyuan
                   and Li, Huixia
                   and Li, Jiashi
                   and Li, Liang
                   and Li, Xiaojie
                   and Li, Xunsong
                   and Li, Yifu
                   and Lin, Shanchuan
                   and Lin, Zhijie
                   and Liu, Jiawei
                   and Liu, Shu
                   and Nie, Xiaonan
                   and Qing, Zhiwu
                   and Ren, Yuxi
                   and Sun, Li
                   and Tian, Zhi
                   and Wang, Rui
                   and Wang, Sen
                   and Wei, Guoqiang
                   and Wu, Guohong
                   and Wu, Jie
                   and Xia, Ruiqi
                   and Xiao, Fei
                   and Xiao, Xuefeng
                   and Yan, Jiangqiao
                   and Yang, Ceyuan
                   and Yang, Jianchao
                   and Yang, Runkai
                   and Yang, Tao
                   and Yang, Yihang
                   and Ye, Zilyu
                   and Zeng, Xuejiao
                   and Zeng, Yan
                   and Zhang, Heng
                   and Zhao, Yang
                   and Zheng, Xiaozheng
                   and Zhu, Peihao
                   and Zou, Jiaxin
                   and Zuo, Feilong},
  journal       = {arXiv preprint arXiv:2506.09113},
  year          = {2025},
  month         = jun,
  eprint        = {2506.09113},
  archivePrefix = {arXiv},
  primaryClass  = {cs.CV},
  doi           = {10.48550/arXiv.2506.09113},
  url           = {https://arxiv.org/abs/2506.09113},
  note          = {Seedance 1.0 Technical Report}
}

@article{teamwan2025wan,
  title         = {{Wan}: Open and Advanced Large-Scale Video Generative Models},
  author        = {{Team Wan}
                   and Wang, Ang
                   and Ai, Baole
                   and Wen, Bin
                   and Mao, Chaojie
                   and Xie, Chen-Wei
                   and Chen, Di
                   and Yu, Feiwu
                   and Zhao, Haiming
                   and Yang, Jianxiao
                   and Zeng, Jianyuan
                   and Wang, Jiayu
                   and Zhang, Jingfeng
                   and Zhou, Jingren
                   and Wang, Jinkai
                   and Chen, Jixuan
                   and Zhu, Kai
                   and Zhao, Kang
                   and Yan, Keyu
                   and Huang, Lianghua
                   and Feng, Mengyang
                   and Zhang, Ningyi
                   and Li, Pandeng
                   and Wu, Pingyu
                   and Chu, Ruihang
                   and Feng, Ruili
                   and Zhang, Shiwei
                   and Sun, Siyang
                   and Fang, Tao
                   and Wang, Tianxing
                   and Gui, Tianyi
                   and Weng, Tingyu
                   and Shen, Tong
                   and Lin, Wei
                   and Wang, Wei
                   and Wang, Wei
                   and Zhou, Wenmeng
                   and Wang, Wente
                   and Shen, Wenting
                   and Yu, Wenyuan
                   and Shi, Xianzhong
                   and Huang, Xiaoming
                   and Xu, Xin
                   and Kou, Yan
                   and Lv, Yangyu
                   and Li, Yifei
                   and Liu, Yijing
                   and Wang, Yiming
                   and Zhang, Yingya
                   and Huang, Yitong
                   and Li, Yong
                   and Wu, You
                   and Liu, Yu
                   and Pan, Yulin
                   and Zheng, Yun
                   and Hong, Yuntao
                   and Shi, Yupeng
                   and Feng, Yutong
                   and Jiang, Zeyinzi
                   and Han, Zhen
                   and Wu, Zhi-Fan
                   and Liu, Ziyu},
  journal       = {arXiv preprint arXiv:2503.20314},
  year          = {2025},
  month         = mar,
  eprint        = {2503.20314},
  archivePrefix = {arXiv},
  primaryClass  = {cs.CV},
  doi           = {10.48550/arXiv.2503.20314},
  url           = {https://arxiv.org/abs/2503.20314}
}

@inproceedings{Assran_2023_CVPR,
  author    = {Assran, Mahmoud and Duval, Quentin and Misra, Ishan
               and Bojanowski, Piotr and Vincent, Pascal
               and Rabbat, Michael and LeCun, Yann and Ballas, Nicolas},
  title     = {Self-Supervised Learning From Images With a Joint-Embedding Predictive Architecture},
  booktitle = {Proceedings of the IEEE/CVF Conference on Computer Vision and Pattern Recognition (CVPR)},
  pages     = {15619--15629},
  month     = jun,
  year      = {2023}
}

@misc{assran2025vjepa2,
  title         = {{V-JEPA 2}: Self-Supervised Video Models Enable Understanding, Prediction and Planning},
  author        = {Assran, Mido
                   and Bardes, Adrien
                   and Fan, David
                   and Garrido, Quentin
                   and Howes, Russell
                   and Komeili, Mojtaba
                   and Muckley, Matthew
                   and Rizvi, Ammar
                   and Roberts, Claire
                   and Sinha, Koustuv
                   and Zholus, Artem
                   and Arnaud, Sergio
                   and Gejji, Abha
                   and Martin, Ada
                   and Hogan, Fran{\c{c}}ois Robert
                   and Dugas, Daniel
                   and Bojanowski, Piotr
                   and Khalidov, Vasil
                   and Labatut, Patrick
                   and Massa, Francisco
                   and Szafraniec, Marc
                   and Krishnakumar, Kapil
                   and Li, Yong
                   and Ma, Xiaodong
                   and Chandar, Sarath
                   and Meier, Franziska
                   and LeCun, Yann
                   and Rabbat, Michael
                   and Ballas, Nicolas},
  year          = {2025},
  month         = jun,
  eprint        = {2506.09985},
  archivePrefix = {arXiv},
  primaryClass  = {cs.AI},
  doi           = {10.48550/arXiv.2506.09985},
  url           = {https://arxiv.org/abs/2506.09985}
}

@article{nvidia2025cosmos,
  title         = {{Cosmos} World Foundation Model Platform for Physical {AI}},
  author        = {{NVIDIA}},
  journal       = {arXiv preprint arXiv:2501.03575},
  year          = {2025},
  month         = jan,
  eprint        = {2501.03575},
  archivePrefix = {arXiv},
  primaryClass  = {cs.CV},
  doi           = {10.48550/arXiv.2501.03575},
  url           = {https://arxiv.org/abs/2501.03575}
}

@article{nvidia2026cosmos3,
  title         = {{Cosmos 3}: Omnimodal World Models for Physical {AI}},
  author        = {{NVIDIA}},
  journal       = {arXiv preprint arXiv:2606.02800},
  year          = {2026},
  month         = jun,
  eprint        = {2606.02800},
  archivePrefix = {arXiv},
  primaryClass  = {cs.CV},
  doi           = {10.48550/arXiv.2606.02800},
  url           = {https://arxiv.org/abs/2606.02800}
}

@inproceedings{vaswani2017attention,
  title     = {Attention Is All You Need},
  author    = {Vaswani, Ashish
               and Shazeer, Noam
               and Parmar, Niki
               and Uszkoreit, Jakob
               and Jones, Llion
               and Gomez, Aidan N.
               and Kaiser, {\L}ukasz
               and Polosukhin, Illia},
  booktitle = {Advances in Neural Information Processing Systems},
  volume    = {30},
  pages     = {5998--6008},
  publisher = {Curran Associates, Inc.},
  year      = {2017},
  url       = {https://proceedings.neurips.cc/paper/7181-attention-is-all-you-need}
}

@inproceedings{ho2020denoising,
  title     = {Denoising Diffusion Probabilistic Models},
  author    = {Ho, Jonathan
               and Jain, Ajay
               and Abbeel, Pieter},
  booktitle = {Advances in Neural Information Processing Systems},
  volume    = {33},
  pages     = {6840--6851},
  publisher = {Curran Associates, Inc.},
  year      = {2020},
  url       = {https://proceedings.neurips.cc/paper/2020/hash/4c5bcfec8584af0d967f1ab10179ca4b-Abstract.html}
}

@inproceedings{Wang_2025_CVPR,
  author    = {Wang, Jianyuan
               and Chen, Minghao
               and Karaev, Nikita
               and Vedaldi, Andrea
               and Rupprecht, Christian
               and Novotny, David},
  title     = {{VGGT}: Visual Geometry Grounded Transformer},
  booktitle = {Proceedings of the IEEE/CVF Conference on Computer Vision and Pattern Recognition (CVPR)},
  pages     = {5294--5306},
  month     = jun,
  year      = {2025},
  url       = {https://openaccess.thecvf.com/content/CVPR2025/html/Wang_VGGT_Visual_Geometry_Grounded_Transformer_CVPR_2025_paper.html}
}

@misc{sam3dteam2025sam3d,
  title         = {{SAM 3D}: 3Dfy Anything in Images},
  author        = {{SAM 3D Team}
                   and Chen, Xingyu
                   and Chu, Fu-Jen
                   and Gleize, Pierre
                   and Liang, Kevin J.
                   and Sax, Alexander
                   and Tang, Hao
                   and Wang, Weiyao
                   and Guo, Michelle
                   and Hardin, Thibaut
                   and Li, Xiang
                   and Lin, Aohan
                   and Liu, Jiawei
                   and Ma, Ziqi
                   and Sagar, Anushka
                   and Song, Bowen
                   and Wang, Xiaodong
                   and Yang, Jianing
                   and Zhang, Bowen
                   and Doll{\'a}r, Piotr
                   and Gkioxari, Georgia
                   and Feiszli, Matt
                   and Malik, Jitendra},
  year          = {2025},
  month         = nov,
  eprint        = {2511.16624},
  archivePrefix = {arXiv},
  primaryClass  = {cs.CV},
  doi           = {10.48550/arXiv.2511.16624},
  url           = {https://arxiv.org/abs/2511.16624}
}

@misc{yang2026sam3dbody,
  title         = {{SAM 3D Body}: Robust Full-Body Human Mesh Recovery},
  author        = {Yang, Xitong
                   and Kukreja, Devansh
                   and Pinkus, Don
                   and Sagar, Anushka
                   and Fan, Taosha
                   and Park, Jinhyung
                   and Shin, Soyong
                   and Cao, Jinkun
                   and Liu, Jiawei
                   and Ugrinovic, Nicolas
                   and Feiszli, Matt
                   and Malik, Jitendra
                   and Doll{\'a}r, Piotr
                   and Kitani, Kris},
  year          = {2026},
  month         = feb,
  eprint        = {2602.15989},
  archivePrefix = {arXiv},
  primaryClass  = {cs.CV},
  doi           = {10.48550/arXiv.2602.15989},
  url           = {https://arxiv.org/abs/2602.15989}
}

@misc{lin2025depthanything3,
  title         = {{Depth Anything 3}: Recovering the Visual Space from Any Views},
  author        = {Lin, Haotong
                   and Chen, Sili
                   and Liew, Junhao
                   and Chen, Donny Y.
                   and Li, Zhenyu
                   and Shi, Guang
                   and Feng, Jiashi
                   and Kang, Bingyi},
  year          = {2025},
  month         = nov,
  eprint        = {2511.10647},
  archivePrefix = {arXiv},
  primaryClass  = {cs.CV},
  doi           = {10.48550/arXiv.2511.10647},
  url           = {https://arxiv.org/abs/2511.10647}
}

@inproceedings{Wang_2026_CVPR,
  author    = {Wang, Jianyuan
               and Chen, Minghao
               and Zhang, Shangzhan
               and Karaev, Nikita
               and Sch{\"o}nberger, Johannes
               and Labatut, Patrick
               and Bojanowski, Piotr
               and Novotny, David
               and Vedaldi, Andrea
               and Rupprecht, Christian},
  title     = {{VGGT}-$\Omega$},
  booktitle = {Proceedings of the IEEE/CVF Conference on Computer Vision and Pattern Recognition (CVPR)},
  pages     = {21486--21499},
  month     = jun,
  year      = {2026},
  url       = {https://openaccess.thecvf.com/content/CVPR2026/html/Wang_VGGT-ohm_CVPR_2026_paper.html}
}

@inproceedings{Xiang_2026_CVPR,
  author    = {Xiang, Jianfeng
               and Chen, Xiaoxue
               and Xu, Sicheng
               and Wang, Ruicheng
               and Lv, Zelong
               and Deng, Yu
               and Zhu, Hongyuan
               and Dong, Yue
               and Zhao, Hao
               and Yuan, Nicholas Jing
               and Yang, Jiaolong},
  title     = {Native and Compact Structured Latents for {3D} Generation},
  booktitle = {Proceedings of the IEEE/CVF Conference on Computer Vision and Pattern Recognition (CVPR)},
  pages     = {14419--14429},
  month     = jun,
  year      = {2026},
  url       = {https://openaccess.thecvf.com/content/CVPR2026/html/Xiang_Native_and_Compact_Structured_Latents_for_3D_Generation_CVPR_2026_paper.html}
}

@article{klingteam2026motioncontrol,
  title         = {{Kling-MotionControl} Technical Report},
  author        = {{Kling Team}
                   and Chen, Jialu
                   and Ding, Yikang
                   and Fang, Zhixue
                   and Gai, Kun
                   and He, Kang
                   and He, Xu
                   and Hua, Jingyun
                   and Lao, Mingming
                   and Li, Xiaohan
                   and Liu, Hui
                   and Liu, Jiwen
                   and Liu, Xiaoqiang
                   and Shi, Fan
                   and Shi, Xiaoyu
                   and Sun, Peiqin
                   and Tang, Songlin
                   and Wan, Pengfei
                   and Wen, Tiancheng
                   and Wu, Zhiyong
                   and Zhang, Haoxian
                   and Zhao, Runze
                   and Zhang, Yuanxing
                   and Zhou, Yan},
  journal       = {arXiv preprint arXiv:2603.03160},
  year          = {2026},
  month         = mar,
  eprint        = {2603.03160},
  archivePrefix = {arXiv},
  primaryClass  = {cs.CV},
  doi           = {10.48550/arXiv.2603.03160},
  url           = {https://arxiv.org/abs/2603.03160}
}

@misc{sensetime2026seko,
  author       = {{SenseTime}},
  title        = {{Seko}: World-Class AI Video Generation Platform},
  year         = {2026},
  howpublished = {Official website},
  url          = {https://seko.sensetime.com/},
  note         = {Accessed: 2026-06-28}
}

@misc{tapnow2026,
  author       = {{TapNow}},
  title        = {{TapNow}: Your Agentic Creative Canvas},
  year         = {2026},
  howpublished = {Official website},
  url          = {https://www.tapnow.ai/},
  note         = {Accessed: 2026-06-28}
}

@misc{lovart2026,
  author       = {{Lovart}},
  title        = {{Lovart}: The World's First AI Design Agent},
  year         = {2026},
  howpublished = {Official website},
  url          = {https://www.lovart.ai/},
  note         = {Accessed: 2026-06-28}
}

@misc{liblibai2026libtv,
  author       = {{LiblibAI}},
  title        = {{LibTV}: Professional Video Creation Platform},
  year         = {2026},
  howpublished = {Official website},
  url          = {https://www.liblib.tv/},
  note         = {Accessed: 2026-06-28}
}

@misc{worldlabs2025marble,
  author       = {{World Labs}},
  title        = {{Marble}: A Multimodal World Model},
  year         = {2025},
  month        = nov,
  howpublished = {Official product announcement},
  url          = {https://www.worldlabs.ai/blog/marble-world-model},
  note         = {Published November 12, 2025; accessed: 2026-06-28}
}

@inproceedings{hu2022lora,
  title     = {{LoRA}: Low-Rank Adaptation of Large Language Models},
  author    = {Hu, Edward J. and
               Shen, Yelong and
               Wallis, Phillip and
               Allen-Zhu, Zeyuan and
               Li, Yuanzhi and
               Wang, Shean and
               Wang, Lu and
               Chen, Weizhu},
  booktitle = {International Conference on Learning Representations},
  year      = {2022},
  url       = {https://openreview.net/forum?id=nZeVKeeFYf9}
}

@inproceedings{zhang2023controlnet,
  title     = {Adding Conditional Control to Text-to-Image Diffusion Models},
  author    = {Zhang, Lvmin and
               Rao, Anyi and
               Agrawala, Maneesh},
  booktitle = {Proceedings of the IEEE/CVF International Conference on Computer Vision},
  pages     = {3836--3847},
  month     = {October},
  year      = {2023},
  url       = {https://openaccess.thecvf.com/content/ICCV2023/html/Zhang_Adding_Conditional_Control_to_Text-to-Image_Diffusion_Models_ICCV_2023_paper.html}
}

@article{tschannen2025siglip2,
  title   = {{SigLIP 2}: Multilingual Vision-Language Encoders with Improved
             Semantic Understanding, Localization, and Dense Features},
  author  = {Tschannen, Michael and
             Gritsenko, Alexey and
             Wang, Xiao and
             Naeem, Muhammad Ferjad and
             Alabdulmohsin, Ibrahim and
             Parthasarathy, Nikhil and
             Evans, Talfan and
             Beyer, Lucas and
             Xia, Ye and
             Mustafa, Basil and
             H{\'e}naff, Olivier and
             Harmsen, Jeremiah and
             Steiner, Andreas and
             Zhai, Xiaohua},
  journal = {arXiv preprint arXiv:2502.14786},
  year    = {2025},
  doi     = {10.48550/arXiv.2502.14786},
  eprint        = {2502.14786},
  archivePrefix = {arXiv},
  primaryClass  = {cs.CV},
  url           = {https://arxiv.org/abs/2502.14786}
}

@article{wang2025worldgen,
  title         = {{WorldGen}: From Text to Traversable and Interactive {3D} Worlds},
  author        = {Wang, Dilin and Jung, Hyunyoung and Monnier, Tom and Sohn, Kihyuk
                   and Zou, Chuhang and Xiang, Xiaoyu and Yeh, Yu-Ying and Liu, Di
                   and Huang, Zixuan and Nguyen-Phuoc, Thu and Fan, Yuchen
                   and Oprea, Sergiu and Wang, Ziyan and Shapovalov, Roman
                   and Sarafianos, Nikolaos and Groueix, Thibault and Toisoul, Antoine
                   and Dhar, Prithviraj and Chu, Xiao and Chen, Minghao
                   and Park, Geon Yeong and Gupta, Mahima and Azziz, Yassir
                   and Ranjan, Rakesh and Vedaldi, Andrea},
  journal       = {arXiv preprint arXiv:2511.16825},
  year          = {2025},
  eprint        = {2511.16825},
  archivePrefix = {arXiv},
  primaryClass  = {cs.CV},
  doi           = {10.48550/arXiv.2511.16825},
  url           = {https://arxiv.org/abs/2511.16825}
}

@techreport{openai2026gpt55,
  author      = {{OpenAI}},
  title       = {{GPT-5.5 System Card}},
  institution = {{OpenAI}},
  year        = {2026},
  month       = apr,
  url         = {https://openai.com/index/gpt-5-5-system-card/}
}

@misc{capcut2026,
  author       = {{CapCut}},
  title        = {{CapCut: AI-Powered Photo and Video Editor for Everyone}},
  year         = {2026},
  howpublished = {\url{https://www.capcut.com/}},
  note         = {Accessed: 2026-06-29}
}

@misc{nanoai2026,
  author       = {{360 Group}},
  title        = {{Nano AI: Your Personal Super Agent}},
  year         = {2026},
  howpublished = {\url{https://www.n.cn/}},
  note         = {Accessed: 2026-06-29}
}

@inproceedings{lee2022autoregressive,
  author    = {Lee, Doyup and Kim, Chiheon and Kim, Saehoon
               and Cho, Minsu and Han, Wook-Shin},
  title     = {Autoregressive Image Generation Using Residual Quantization},
  booktitle = {Proceedings of the IEEE/CVF Conference on Computer Vision
               and Pattern Recognition},
  pages     = {11523--11532},
  month     = jun,
  year      = {2022}
}

@misc{openai2025codex,
  author       = {{OpenAI}},
  title        = {Introducing {Codex}},
  year         = {2025},
  month        = may,
  howpublished = {\url{https://openai.com/index/introducing-codex/}}
}

@article{inspatio2026world,
  author        = {{InSpatio Team}
                   and Shen, Donghui
                   and Zhang, Guofeng
                   and Liu, Haomin
                   and Ji, Haoyu
                   and Bao, Hujun
                   and Zhai, Hongjia
                   and Liu, Jialin
                   and Guo, Jing
                   and Wang, Nan
                   and Pan, Siji
                   and Pan, Weihong
                   and Xie, Weijian
                   and Liu, Xianbin
                   and Xiang, Xiaojun
                   and Zhang, Xiaoyu
                   and Chen, Xinyu
                   and Wang, Yifu
                   and Chen, Yipeng
                   and Fan, Zhenzhou
                   and Le, Zhewen
                   and Ye, Zhichao
                   and Zhao, Ziqiang},
  title         = {{INSPATIO-WORLD}: A Real-Time {4D} World Simulator
                   via Spatiotemporal Autoregressive Modeling},
  journal       = {arXiv preprint arXiv:2604.07209},
  year          = {2026},
  eprint        = {2604.07209},
  archivePrefix = {arXiv},
  primaryClass  = {cs.CV},
  doi           = {10.48550/arXiv.2604.07209},
  url           = {https://arxiv.org/abs/2604.07209}
}
\newpage

\end{document}